\documentclass[final,5p,times,twocolumn]{elsarticle}

\usepackage{amssymb}
\usepackage{amsfonts}
\usepackage{amsthm}
\usepackage{graphics} 
\usepackage{epsfig} 
\usepackage{amsmath} 
\usepackage{subfigure}
\usepackage{multirow}
\usepackage{makecell}
\usepackage{graphicx}
\usepackage{lscape}
\usepackage{multicol} 
\usepackage{graphicx, epstopdf}
\usepackage[table,xcdraw]{xcolor}
\usepackage{caption}
\usepackage{xspace,epsfig,url}
\usepackage{enumerate}
\usepackage{float}
\usepackage{epsf}
\usepackage{psfrag}
\usepackage{verbatim}
\usepackage[all]{xy}
\usepackage{color,soul}
\usepackage{bm}
\usepackage{comment}
\usepackage{cases}
\usepackage{hyperref}
\usepackage{physics}
\usepackage{amssymb}
\usepackage{pifont}
\usepackage{array} 
\usepackage[ruled,vlined,linesnumbered]{algorithm2e}
\usepackage{textcomp}
\usepackage{stfloats}
\usepackage{caption}
\usepackage{nicefrac, xfrac}
\usepackage{mathtools}

\newcommand{\expnumber}[2]{{#1}\mathrm{e}{#2}}

\journal{ArXiv Preprint}

\begin{document}

\begin{frontmatter}



\title{Multimodal Optimization with k-Cluster Big Bang-Big Crunch Algorithm and Postprocessing Methods for Identification and Quantification of Optima}


\author[inst1]{Kemal Erdem Yenin}
\author[inst1]{Reha Oguz Sayin}
\author[inst1]{Kuzey Arar}
\author[inst1]{Kadir Kaan Atalay}
\author[inst1]{Fabio Stroppa}

\affiliation[inst1]{organization={Computer Engineering Department, Kadir Has University},
            addressline={Cibali, Kadir Has Cd., Fatih}, 
            city={Istanbul},
            postcode={34083}, 
            country={Turkey}}

\begin{abstract}
Multimodal optimization is often encountered in engineering problems, especially when different and alternative solutions are sought. Evolutionary algorithms can efficiently tackle multimodal optimization thanks to their features such as the concept of population, exploration/exploitation, and being suitable for parallel computation. 
This paper investigates whether a less-known optimizer, the Big Bang-Big Crunch (BBBC) algorithm, is suitable for multimodal optimization. We extended BBBC and propose k-BBBC, a clustering-based multi-modal optimizer. Additionally, we introduce two post-processing methods to (i) identify the local optima in a set of retrieved solutions (i.e., a population), and (ii) quantify the number of correctly retrieved optima against the expected ones (i.e., success rate). 
Our results show that k-BBBC performs well even with problems having a large number of optima (tested on $379$ optima) and high dimensionality (tested on $32$ decision variables), but it becomes computationally too expensive for problems with many local optima (i.e., in the CEC'2013 benchmark set). Compared to other multimodal optimization methods, it outperforms them in terms of accuracy (in both search and objective space) and success rate (number of correctly retrieved optima) when tested on basic multimodal functions, especially when elitism is applied; however, it requires knowing the number of optima of a problem, which makes its performance decrease when tested on niching competition test CEC'2013. Lastly, we validated our proposed post-processing methods by comparing their success rate to the actual one: results suggest that these methods can be used to evaluate the performance of a multimodal optimization algorithm by correctly identifying optima and providing an indication of success -- without the need to know where the optima are located in the search space.
\end{abstract}


\begin{keyword}
Multimodal Optimization \sep Evolutionary Algorithms \sep Clustering, Big Bang-Big Crunch
\PACS 	87.85.St \sep 	87.55.de \sep 87.55.kd
\MSC 49 
\end{keyword}

\end{frontmatter}


\section{Introduction}
\label{sec:intro}


Mathematical optimization is the search for the best element within a set of alternatives based on specific criteria, with applications that range in many different areas such as science, economics, and engineering~\cite{stroppa2023optimizing, stroppa2024optimizing, pardalos2000recent,sioshansi2017optimization,statnikov2012multicriteria,andersson2000survey}.
The process usually consists of exploring the search space of a given objective function, which could be limited by well-defined constraints, to retrieve a single global optimal solution that either minimizes or maximizes the objective -- namely, global optimization. When the objective function features more than one suitable solution (i.e., optima or modes), the process is known as multimodal optimization. 

Many practical optimization problems have more than one optimal solution -- whether that is multiple optima with the same objective value or a single global optimum surrounded by many local optima. In such cases, retrieving a single optimum might not be sufficient to tackle the problem due to different reasons:
    (i) a global optimum might be defined by parameters that are unrealistic or excessively expensive~\cite{cuevas2014cuckoo};
    (ii) a solution can be optimal based on the availability of essential resources or the fulfillment of specific criteria, but if these conditions change over time, users need to adapt to different solutions~\cite{deb2012multimodal}; 
    (iii) knowing various ideal solutions within the search space can offer a valuable understanding of the characteristics of the problem and its solutions~\cite{deb2008innovization}; 
    (iv) robust solutions are preferred over global ones because if the optimal solution lies on a peak of the function, any small change in its value would cause instability to the system, whereas a sub-optimal solution lying in a wider area or a plateau is less susceptible to change~\cite{nomaguchi2016robust}; and
    (v) reliable solutions are preferred over global ones, as uncertainty in the search space might lead to infeasible optimal solutions (i.e., in violation of constraints)~\cite{dizangian2015reliability}.
Under these circumstances, a local optimum offering decent performance at a reasonable cost might be more desirable than a costly global optimum providing a marginal better performance~\cite{wong2012evolutionary}.

Evolutionary Computation offers a set of optimization algorithms, namely Evolutionary Algorithms (EAs), which can easily tackle complex optimization problems such as multimodal ones. Thanks to their stochastic exploration of space, along with the inherent parallel computation offered by their population nature, they can easily locate different good solutions at the same time~\cite{goldberg1989genetic,holland1975adaptation}. 
Many multimodal evolutionary algorithms (MMEAs) have been proposed in the literature, featuring diverse strategies such as: 
(i) implicitly promote diversity by spatial segregation~\cite{izzo2012generalized,dick2003spatially} or spatial distribution~\cite{husbands1994distributed,white1997double,alba2008introduction,gordon1999terrain,gordon2004visualization,llora2001knowledge} of a single population;
(ii) partitioning the population~\cite{kashtiban2016solving} and imposing mating restrictions between different populations~\cite{ursem1999multinational,thomsen2000religion} or depending on the individuals' life cycle~\cite{kubota1994genetic};
(iii) with elitist methods~\cite{liang2011genetic} or methods that conserve the genotype of individuals~\cite{li2002species}; 
(iv) enforcing diversity with fitness sharing~\cite{goldberg1987genetic, beasley1993sequential, miller1996genetic}, or clearing~\cite{petrowski1996clearing}, crowding~\cite{dejong1975analysis,mahfoud1992crowding,mahfoud1993simple,mahfoud1995niching,mengshoel1999probabilistic,thomsen2004multimodal}, clustering~\cite{yin1993fast}; and
(v) employing multi-objective optimization techniques~\cite{deb2012multimodal}.
However, most of these methods suffer from difficulties such as tuning niching parameters, poor scalability with high dimensionality, extra computational overhead, not being able to fully explore the search space, relying on gradient information, premature convergence, and failure to find the right balance between exploration (finding new solutions in the search space) and exploitation (making use of the features of the good solutions discovered throughout the run)~\cite{chen2009preserving,hong2022balancing}.

In this work, we investigate how the efficient and yet relatively unknown Big Bang-Big Crunch (BBBC) algorithm~\cite{erol2006new} can be extended to solve multimodal optimization problems. This investigation is motivated by the advantages of BBBC over genetic algorithms: premature convergence, convergence speed, and execution time. Our proposed algorithm, namely the k-Cluster Big Bang-Big Crunch algorithm (k-BBBC), is a method that relies on uniform exploration of the search space, as well as on exploitation of its best solutions -- with the downside of knowing an estimation of the number of optima sought. An additional strength of BBBC is to dynamically turn exploration into exploitation during a single run, guaranteeing convergence to an optimal set of solutions; we used this property, allowing k-BBBC to correctly and fully converge to the local optima of a problem -- note that the entire population will converge. However, the amount of initial exploration must be sufficient to find all the optima~\cite{roy2013simulated, yahyaie2011surrogate}. To ensure a proper amount of exploration, we mathematically defined the relationship between the population size and the number of local optima to be retrieved (when known), which allowed k-BBBC to find all optima for every problem we tested -- with some minor exceptions for basic multimodal functions, and for large-scale problems that require more advanced techniques. 

However, returning a population of converged solutions without identifying which ones are the optima is not efficient (i.e., we cannot simply sort solutions based on their fitness). We certainly do not want to manually compare solutions against each other: this quickly becomes unfeasible when the population size and the number of expected optima increases. Therefore, we propose efficient and automated methods for post-processing the output of an MMEA -- theoretically, every MMEA provided that its population fully converged. As additional contributions, we propose the following two clustering-based methods:

\begin{itemize}
    \item an identification procedure to detect the local optima in a population of converged solutions, and
    \item a quantification procedure to determine how many local optima have been missed (when this information is available).
\end{itemize}  


The rest of the paper is organized as follows.
Sec.~\ref{sec:methods} describes k-BBBC and 
Sec.~\ref{sec:postprocessing} describes post-processing procedures for precise localization of optimal solutions within a wide population of individuals. Then, we performed two experiments. The first, in  
Sec.~\ref{sec:experiments_local}, tests k-BBBC on basic multimodal functions, with dimensionality ranging from 1 to 32, showing that the algorithm correctly converges (in most cases) to all local optima; in this experiment, k-BBBC was compared with the only publicly available MMEA on Mathworks File Exchange, MCS~\cite{cuevas2014cuckoo}, outperforming it. In this section, we also prove that the identification and quantification methods work during post-processing by comparing their results with the known solutions.
The second experiment, in Sec.~\ref{sec:experiments_global},
compares k-BBBC to thirteen state-of-the-art MMEAs developed in the last decade on the CEC'2013 benchmark set~\cite{li2013benchmark}:
this benchmark, however, is designed for algorithms that retrieve multiple global optima only and ignore any local optima -- which are considered as \textit{traps}. Its functions mostly feature a large and unknown number of local minima, which cannot be estimated, making k-BBBC unsuitable for it -- i.e., k-BBBC needs an estimation of the number of optima sought. Nevertheless, the results are surprisingly acceptable when compared with the latest state-of-the-art, considering that the algorithm was not designed for this purpose. Sec.~\ref{sec:discussion} discusses the advantages and disadvantages of k-BBBC based on the experimental results, leaving room for several future works and opportunities that will be discussed in Sec.~\ref{sec:conclusion}. 

\section{Optimization Method}
\label{sec:methods}

This section will present our proposed multimodal optimization method, starting from its original version for global optimization to a full walk-through of our proposed method and its behavior. We will be using the function defined in Eqn.~(\ref{eq:key}), namely the \textit{Key} function, to provide visual reference while describing the algorithm. This function features a single decision variable $x$ and a changeable number of optima depending on the parameter $m$ -- featuring only one global minimum. Lastly, Table~\ref{tab:parameters} reports the list of the algorithm's parameters along with the symbol we will use to refer to them in the following sections.

\begin{equation}
    \label{eq:key}
    \begin{aligned}
    \textrm{minimize}  \quad & f(x,m) = {10 (1+\cos{(2 \pi m x)}) + 2 m x^2} \\
    \textrm{subject to}         \quad & 0 \leq x \leq 1 \\
    \end{aligned}
\end{equation}

\begin{table}[h!]
\centering\caption{Algorithm's Parameters}
    \label{tab:parameters}
    \begin{center}
    \begin{tabular}{|c|c|}
    \hline
    \rowcolor[HTML]{C0C0C0}
    \textbf{Symbol}      & \textbf{Description} \\
    \hline
    \rowcolor[HTML]{EFEFEF} 
    $n$      & population size \\
    \rowcolor[HTML]{FFFFFF}
    $P$ & set of individuals (population) \\ 
    \rowcolor[HTML]{EFEFEF}
    $C$ & set of centers of mass \\ 
    \rowcolor[HTML]{FFFFFF} 
    $g$ & max number of iterations  \\
    \rowcolor[HTML]{EFEFEF} 
    $m$      & number of local optima \\
    \rowcolor[HTML]{FFFFFF}
    $k$ & number of clusters \\ 
    \rowcolor[HTML]{EFEFEF} 
    $f$      & objective function \\ 
    \rowcolor[HTML]{FFFFFF} 
    $d$      & problem dimensionality \\ 
    \hline
    \end{tabular}
    \end{center}
\end{table}


\subsection{Original Big Bang-Big Crunch Algorithm}
\label{sec:bbbc}

The Big Bang-Big Crunch (BBBC) algorithm was proposed by Erol et al.~\cite{erol2006new, gencc2010big} to address some of the biggest disadvantages of genetic algorithms: premature convergence, convergence speed, and execution time. BBBC is inspired by the evolution of the universe, and it is composed of two phases: (i) explosion, or \textit{big bang}, in which the energy dissipation produces disorder and randomness, and (ii) implosion, or \textit{big crunch}, in which randomness is drawn back into a different order. Just like any other EA, BBBC creates an initial random population $P$ uniformly spread throughout the search space (bang), evaluates its individuals, and gathers them into their center of mass (crunch). These two phases are iterated, spreading new solutions closer to the center of mass as the number of generations increases. At the end of the procedure, the center of mass converges to the optimal solution of the problem. 

BBBC features different \textit{crunch} operators~\cite{gencc2010big}, which determine how the center of mass $c$ is calculated. It can be done by weighting the individuals with corresponding fitness evaluations
 or by choosing the most fitting individual as in Eqn.~(\ref{eq:centerOfMass_best}). 


\begin{equation}
    \label{eq:centerOfMass_best}
    \begin{aligned}
        c = 
        \begin{dcases}
            x_j : f(x_j) = \max_{i=1}^{n}{f(x_i)} \; \text{if maximization}\\
            x_j : f(x_j) = \min_{i=1}^{n}{f(x_i)} \; \text{if minimization}
        \end{dcases}, \; x_i \in P\\
    \end{aligned}
\end{equation}

The state-of-the-art reports only one type of \textit{bang} operator~\cite{erol2006new}, which generates new individuals $x^{new}$ for the next population as expressed in Eqn.~(\ref{eq:bang}). The center of mass $c$ is noised with a random number $r$, which is multiplied by the search space bounds $x^{U}$ and $x^{L}$ (to ensure the newly generated individual being within the search space) and divided by the current iteration step $i$ (to progressively reduce the extent of the expansion in the search space and guaranteeing convergence). Note that the iteration step can also be multiplied by a further parameter to adjust the explosion strength and quicken convergence~\cite{genc2013big}.

\begin{equation}
    \label{eq:bang}
    \begin{aligned}
        x^{new} = c + \frac{(x^{U}-x^{L})\cdot r}{i}  , \quad x \in P
    \end{aligned}
\end{equation}

\subsection{Clustering for Multimodal Optimization}
\label{sec:kbbbc}

The metaheuristic of reducing individuals to their center of mass makes BBBC the perfect candidate for being a multimodal optimization solver based on clustering~\cite{yin1993fast}. The hypothesis is that if a single run of BBBC can crunch to the global optimum, then multiple parallel runs on separate regions of the search space can crunch to every local optimum by forming clusters of solutions. 

\begin{figure*}[t!]
    \centering
    \subfigure[\protect\url{}\label{fig:kbbbc_example_early}Early]%
    {\includegraphics[width=5cm]{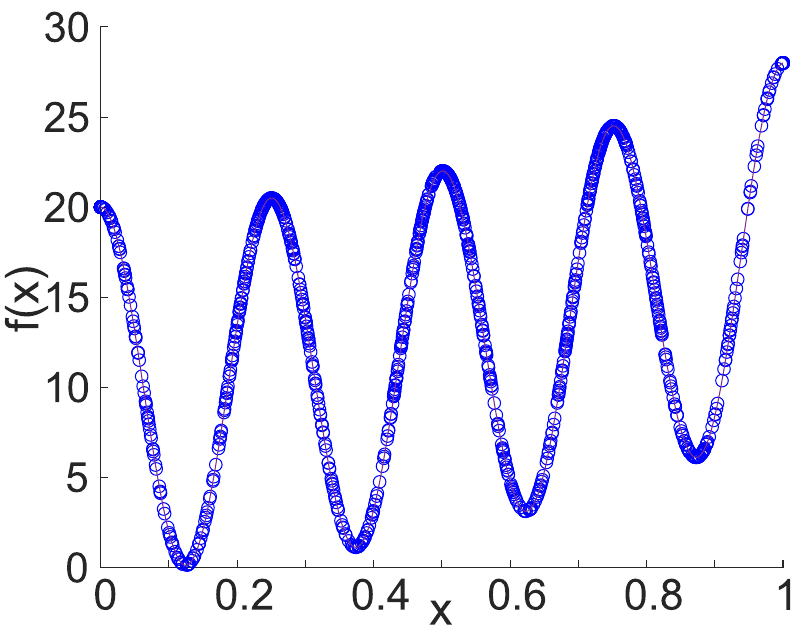}}
    \subfigure[\protect\url{}\label{fig:kbbbc_example_mid}Mid]%
    {\includegraphics[width=5cm]{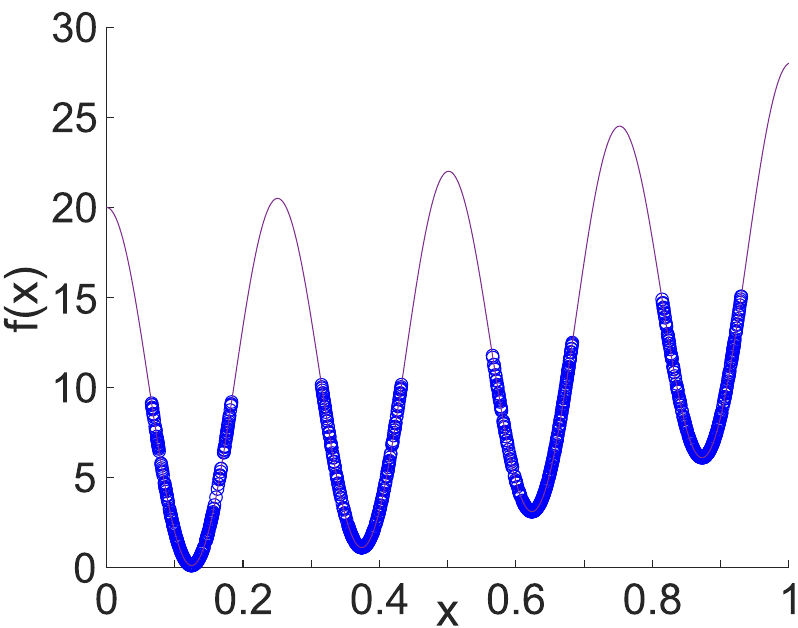}}
    \subfigure[\protect\url{}\label{fig:kbbbc_example_late}Late]%
    {\includegraphics[width=5cm]{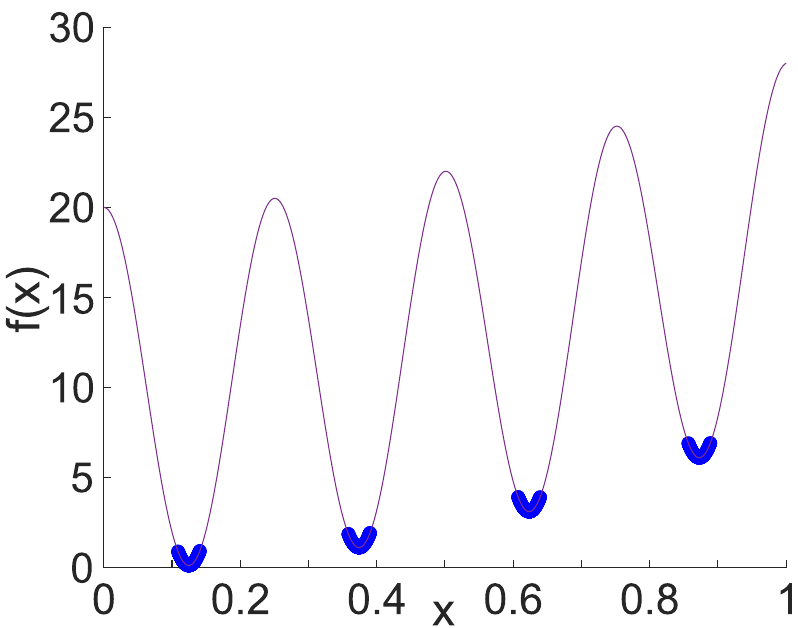}}
    \caption{Typical behavior of k-BBBC on an optimization problem with four local minima. In early generations (a), we can observe a quasi-random population distribution. In mid generations (b), four independent clusters begin to form. Finally, in late generations (c), the entire population correctly converged to the four local minima.}
    \label{fig:kbbbc_example}
\end{figure*}

Based on these considerations, we propose the k-Cluster Big Bang-Big Crunch algorithm (k-BBBC), which is detailed in Algorithm~\ref{alg:kbbbc}. The algorithm starts by generating a population of $n$ random individuals (line 2), which at each iteration (line 3) go through the following steps:

\begin{itemize}
    \item each individual in the population is evaluated and assigned a fitness based on the problem to be optimized (line 6);
    \item the population is divided into $k$ clusters with k-means algorithm~\cite{ahmed2020k}, and each individual is assigned to a cluster (line 7);
    \item each sub-population defined by a cluster is crunched to its center of mass (line 8) -- although k-means returns the set of centers of mass evaluated during clustering, for k-BBBC we used the crunch operator expressed in Eqn.~(\ref{eq:centerOfMass_best}), which selects the best-fitting individual per cluster; and
    \item in the following iteration, each center of mass is expanded to generate a new population of individuals around the $k$ centers of mass (line 5) -- the bang operation is performed as in Eqn.~(\ref{eq:bang}).
\end{itemize}
This process is repeated for $g$ iterations, at the end of which it will return an array of $k$ centers of mass representing the best individuals -- within their region -- of the search space.

\IncMargin{1em}
\begin{algorithm}
        \SetKwData{P}{P}
        \SetKwData{X}{M}
        \SetKwData{opt}{m}
        \SetKwData{Q}{Q}
        \SetKwData{G}{g}
        \SetKwData{K}{k}
        \SetKwData{N}{n}
        \SetKwData{CM}{C}
	\SetKwFunction{Init}{randomInitialization}
        \SetKwFunction{Eval}{evaluation}
        \SetKwFunction{Bang}{bigBang}
        \SetKwFunction{Crunch}{bigCrunch}
        \SetKwFunction{KM}{kMeans}
	\SetKwInOut{Input}{input}
	\SetKwInOut{Output}{output}
        \Input{Number of individuals $\N$, number of generations $\G$, number of clusters $\K$, function to evaluate $f$.}
	\Output{Array of centers of mass $\CM$.}
	\BlankLine		
	
	\Begin{	
    	$\P \leftarrow \Init(\N)$\;
            \For{$i \in [1, \G]$}
            {
                \If{$i \neq 1$}{
                    $\P \leftarrow \Bang(\CM,i)$\;
                }
                $\P \leftarrow \Eval(f,\P)$\;
                $\P \leftarrow \KM(\P,\K)$\;
                $\CM \leftarrow \Crunch(\P)$\;
                
            }
		\KwRet{$\CM$};
	}
	
	\caption{k-BBBC}\label{alg:kbbbc}
\end{algorithm}\DecMargin{1em}

Figure~\ref{fig:kbbbc_example} shows a typical behavior of k-BBBC through three generations (early, mid, and late) on a function with four local minima. The figure does not indicate which individual belongs to which cluster, as it is irrelevant to the algorithm; however, at the end of the run, all points visibly converged to every local minimum -- regardless of their cluster.

\subsection{Parameter Setting and Unexpected Behaviors}
\label{sec:settings}

Knowing the exact number $m$ of local optima for a specific function might not be always possible. In a generic case, k-BBBC will blindly work and locate some of the local optima of the problem, without knowing exactly how many have been missed. However, when the number of local optima is known, it is possible to elaborate different strategies to help k-BBBC converge to the right number of solutions.

\begin{figure*}[t]
    \centering
    \subfigure[\protect\url{}\label{fig:kmeans_err_g24}Correct]%
    {\includegraphics[width=5cm]{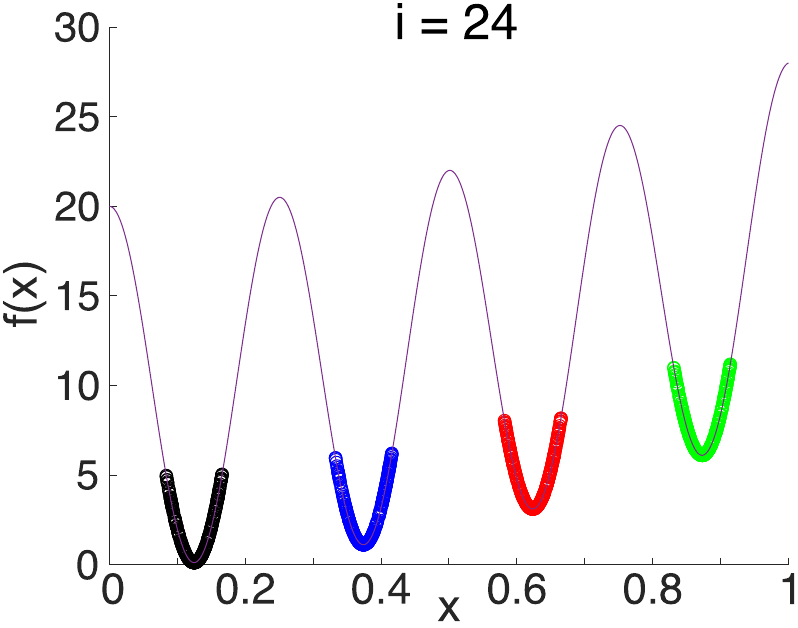}}
    \subfigure[\protect\url{}\label{fig:kmeans_err_g25}Failure]%
    {\includegraphics[width=5cm]{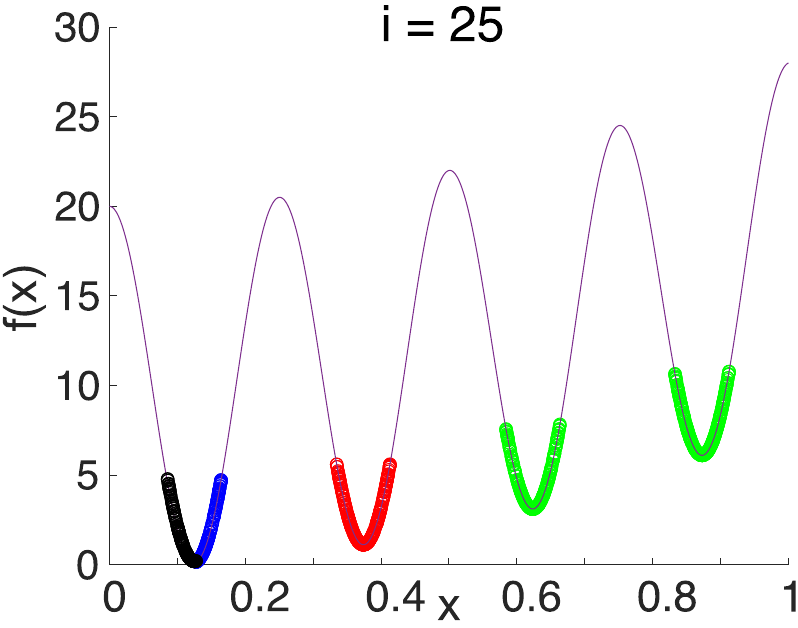}}
    \subfigure[\protect\url{}\label{fig:kmeans_err_g26}Loss]%
    {\includegraphics[width=5cm]{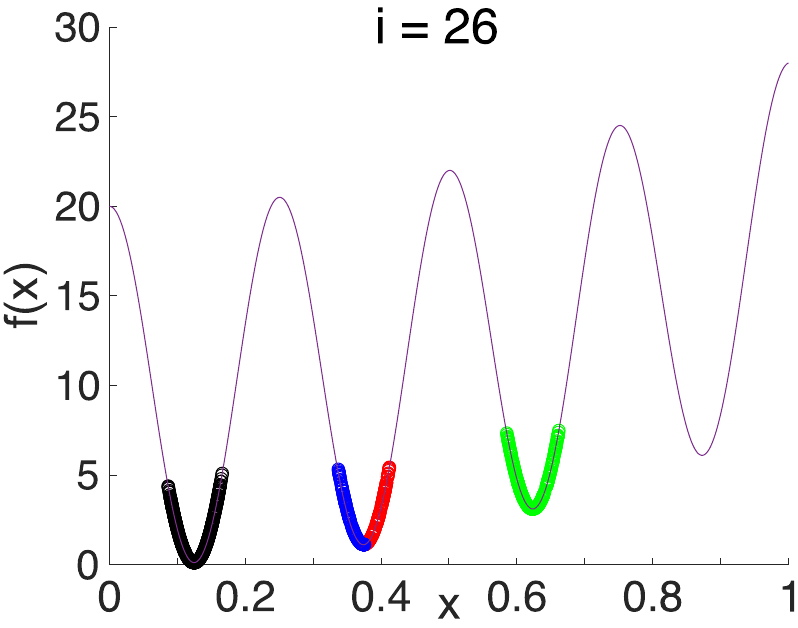}}
    \caption{Example of k-means failure while running k-BBBC on a minimization problem with four local minima, depicting generations 24, 25, and 26. Different colors depict different clusters based on k-means results -- note that the colors do not indicate any correlation between clusters in different generations. 
    In the generation shown in (a), the algorithm correctly detects four clusters after k-BBBC crunches the four sub-populations into their respective optimal values and expands them. In the next generation (b), however, k-means erroneously divides the leftmost cluster ($x = 0.12$) into two, which causes the two rightmost clusters ($x = 0.62, 0.87$) to be merged -- or possibly the other way around. Finally, in the following generation (c), the points belonging to the rightmost merged cluster in (b) crunched to their best value ($x = 0.62$), and when the sub-population expands, the cluster belonging to the rightmost optimum ($x = 0.87$) is lost. Additionally, one random cluster must be split into two ($x = 0.37$), as k-means is still searching for four clusters in a dataset with only three. Unfortunately, this behavior is also observed with k-medoids.}
    \label{fig:kmeans_error}
\end{figure*}

During development, we observed an unexpected outcome from k-means: sometimes, a well-defined set of points is erroneously separated into two clusters, resulting in two spatially-separated sets of points to be merged in a single cluster as k-means is forced to divide the dataset into exactly `$k$' clusters (a graphical description is depicted in Figure~\ref{fig:kmeans_error}). This might happen because k-means cannot adapt to different cluster densities, even when the clusters are spherical, have equal radii, or are well-separated~\cite{raykov2016what}.
We observed that over 100 runs of k-BBBC on the function depicted in Figure~\ref{fig:kmeans_error} (which is very simple), k-means has a $61\%$ failure rate, resulting in k-BBBC retrieving only three of the four local optima six times over ten. We also tested an alternative clustering method, k-medoids~\cite{park2009simple}, which differs from k-means in terms of the points chosen to be the centroids of the clusters (in this case, namely the medoids): k-means takes the mean of the dataset, whereas k-medoids uses actual points from the dataset~\cite{madhulatha2011comparison}. Although we observed a lower failure rate on the same settings ($21\%$), k-medoids has a higher computational complexity and does not solve the problem completely.

Therefore, instead of expecting each $k$ cluster to converge exactly to their respective local optima, k-BBBC works with $k > m$. This leads the population to spread in different locations of the search space, eventually converging around every local optimum -- note that this number can change based on the objective function. Such a procedure forces k-BBBC to exploit good solutions found in every partition of the search space (thus combining exploration and exploitation), allowing it to solve optimization problems featuring optimal solutions that are not uniformly distributed in the search space. 

\begin{figure*}[h]
    \centering
    \subfigure[\protect\url{}\label{fig:relationKM_under}$k = m$]%
    {\includegraphics[width=5.5cm]{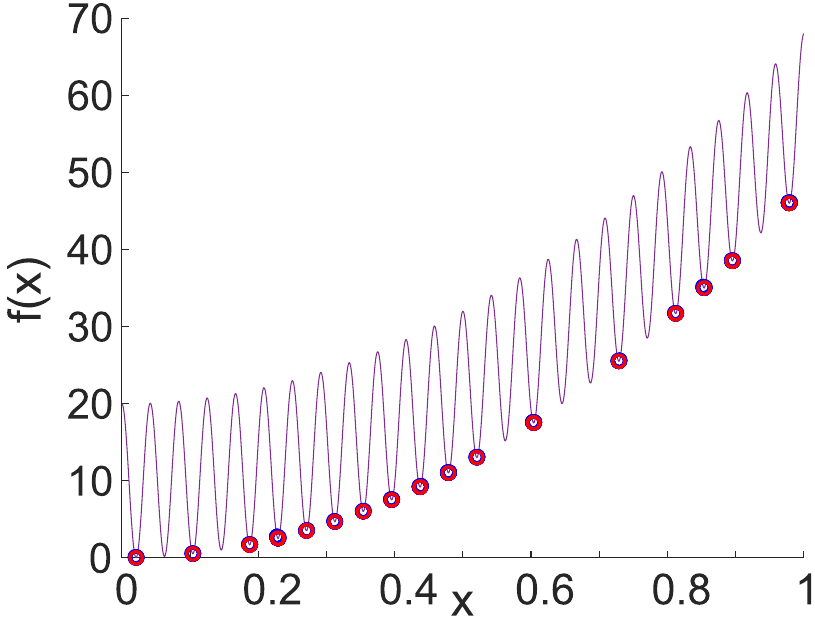}}
    \subfigure[\protect\url{}\label{fig:relationKM_over}$k=5m$]%
    {\includegraphics[width=5.5cm]{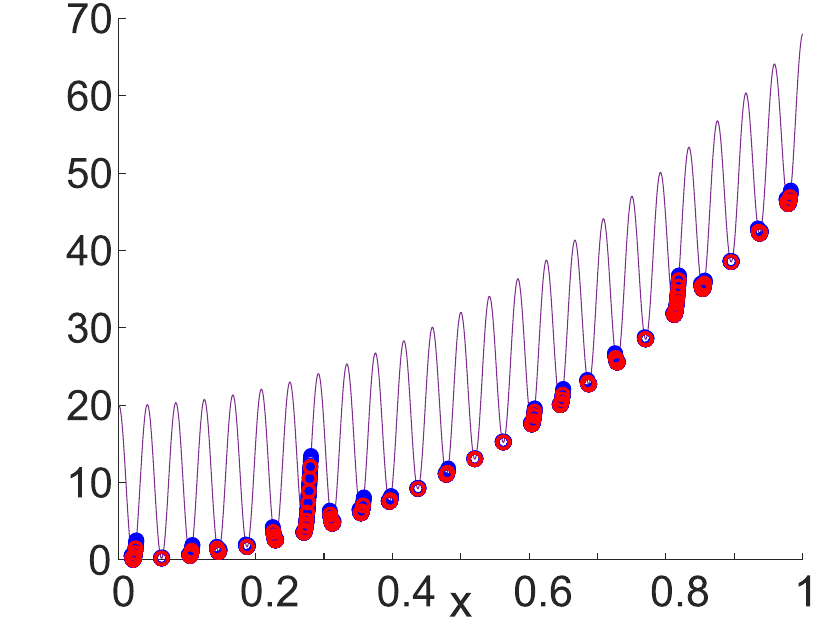}}
    \subfigure[\protect\url{}\label{fig:relationKM_correct}$k=2m$]%
    {\includegraphics[width=5.5cm]{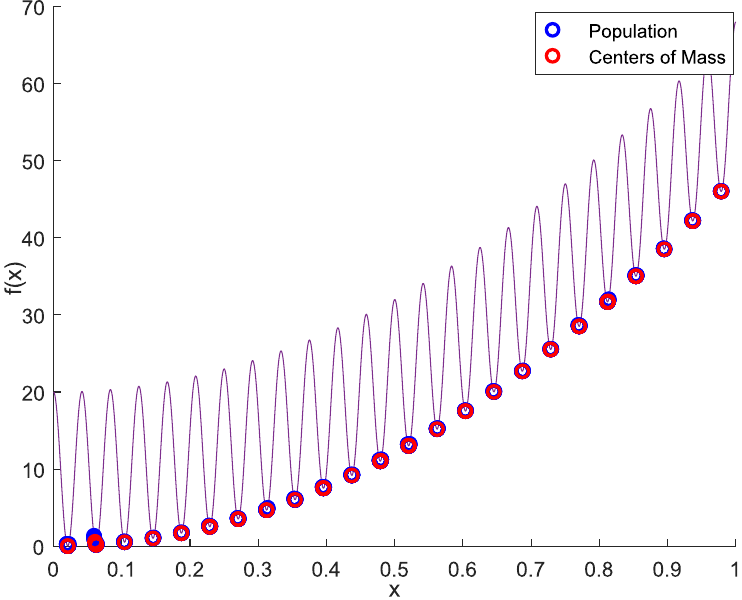}}
    \caption{When setting the value of $k$, it is suggested to divide each concave upward/downward region among at most two clusters per dimension of the problem. When $k$ is underestimated, some optima can be lost; when overestimated, the post-processing methods can be deceived (Sec~\ref{sec:postprocessing}). In this example, we ran k-BBBC on a problem with $m=24$ local minima, $g=1000$ generations, and a population of $n = 20 \cdot k$ individuals.
    In (a), an example of underestimation with $k= m = 24$ producing errors similar to the one reported in Figure~\ref{fig:kmeans_error}, resulting in missing optima.
    In (b), an example of overestimation with $k=5\cdot m = 120$: this results in the population not being completely flattened at the end of the run, and since the population distribution is not uniform through iterations, some clusters will necessarily share part of the search space. 
    In (c), a correct estimation with $k=2 \cdot m = 48$. The centers of mass correctly flattened at the local minima, with repetitions barely visible due to the close proximities of centers of mass within the same concave upward region.}
    \label{fig:relationKM}
\end{figure*}

To prevent uncontrolled outcomes produced by user-defined parameters, we offer a formalization for the relationship between $k$ and $m$. We found that the right balance is to account for two clusters per dimension for each local optimum of the problem, setting the value of $k$ as defined in Eqn.~(\ref{eq:k}).
As shown in Figure~\ref{fig:relationKM}, this relationship provides the right balance to obtain centers of masses crunched in the very proximity of their respective local optimum. By setting $k<m$, some optima can be missed (due to the effect reported in Figure~\ref{fig:kmeans_error}); on the other hand, setting $k \gg m$ will cause the accumulation of centers of mass in a single concave region due to overpopulation, deceiving the post-processing methods described in Sec.~\ref{sec:postprocessing}.

\begin{equation}
    \label{eq:k}
    \begin{aligned}
        k=2 \cdot m \cdot d
    \end{aligned}
\end{equation}

Lastly, we defined a method for determining the population size $n$ depending on the problem to be optimized. In an EA, the population size plays a pivotal role in determining its ability to explore the search space (diversity) and the time it takes to find a satisfactory solution (convergence speed). Adjusting this value is crucial to achieve a balance between solution quality and computational efficiency. Since k-BBBC works by dividing the population into different clusters, the population size must be a multiple of the number of clusters $k$; if the population size is arbitrarily chosen by the user, there is the risk that clusters might not be sufficiently populated. We empirically estimated a multiplication factor for determining $n$ from $k$, starting from small values (i.e., $5$) and progressively increasing it. We found that a value of $20$ was sufficient to explore the search spaces of the tested functions and to correctly converge to local optima. Therefore, the expression in Eqn.~(\ref{eq:n}) allows the population size to vary based on the problem to be optimized, as it depends on the number of clusters and the number of optima. Higher values than $20$ might also be used to increase exploration to the detriment of runtime.

\begin{equation}
    \label{eq:n}
    \begin{aligned}
        n=20 \cdot k
    \end{aligned}
\end{equation}

\subsection{Elitism}
\label{sec:elitism}

Elitism is a strategy in EAs where the best one or more solutions (elites), in each generation,
are inserted into the next without undergoing any change~\cite{ahn2003elitism}. This strategy usually speeds up the convergence of the algorithm~\cite{rudolph1996convergence}.

We propose an elitist version of k-BBBC, which at each iteration reintroduces the best individuals of the past generation into the new one before clustering and crunching. 
The procedure is detailed in Algorithm~\ref{alg:ekbbbc}. In the context of k-BBBC, the elites are the centers of mass retrieved by the crunch operation (line 9), because they are the best individuals in their respective clusters. These $k$ elites -- one per cluster -- will be reintroduced into the population (line 6) after the next pool of individuals is generated with the bang operator (line 5): the new population $P$ will be then composed of $n$ new individuals and $k$ elites. Note that this will not linearly increase the population size at each iteration because the bang operator will always generate a population of size $n$. 

The reason why the population from lines 7 to 9 has a size of $n+k$ rather than always keeping a fixed size of $n$ is that k-means generates new clusters at each generation which are independent of older ones. There is no direct correlation between the elites of the old generation and the bad solutions of the new one because it is not possible to track clusters in consecutive generations; therefore, k-BBBC cannot perform a direct replacement. Note that there is no need to re-evaluate the elites in line 7 because their fitness is already calculated in the previous generation. 

It is also possible to introduce more than one elite per cluster if desired; this, however, would increase the size of the population to be evaluated, clusterized, and then crunched. All to the detriment of runtime.

\IncMargin{1em}
\begin{algorithm}
        \SetKwData{P}{P}
        \SetKwData{X}{M}
        \SetKwData{opt}{m}
        \SetKwData{Q}{Q}
        \SetKwData{G}{g}
        \SetKwData{K}{k}
        \SetKwData{N}{n}
        \SetKwData{CM}{C}
	\SetKwFunction{Init}{randomInitialization}
        \SetKwFunction{Eval}{evaluation}
        \SetKwFunction{Bang}{bigBang}
        \SetKwFunction{Crunch}{bigCrunch}
        \SetKwFunction{KM}{kMeans}
        \SetKwFunction{Elit}{addElites}
	\SetKwInOut{Input}{input}
	\SetKwInOut{Output}{output}
        \Input{Number of individuals $\N$, number of generations $\G$, number of clusters $\K$, function to evaluate $f$.}
	\Output{Array of centers of mass $\CM$.}
	\BlankLine		
	
	\Begin{	
    	$\P \leftarrow \Init(\N)$\;
            \For{$i \in [1, \G]$}
            {
                \If{$i \neq 1$}{
                    $\P \leftarrow \Bang(\CM,i)$\;
                    $\P \leftarrow \P + \CM$\;
                }
                $\P \leftarrow \Eval(f,\P)$\;
                $\P \leftarrow \KM(\P,\K)$\;
                $\CM \leftarrow \Crunch(\P)$\;
                
            }
		\KwRet{$\CM$};
	}
	
	\caption{Elitist k-BBBC}\label{alg:ekbbbc}
\end{algorithm}\DecMargin{1em}

\subsection{Time Complexity}
\label{sec:complexity}

We report the computational time complexity analysis of k-BBBC.
As the population is divided into $k$ clusters, each cluster must perform a number of operations at least linear with the number of its individuals. However, since k-means is a stochastic algorithm, we cannot estimate the correct number of individuals per cluster that k-BBBC will feature throughout its run. Therefore, to evaluate the complexity of k-BBBC we will assume that each cluster contains a fixed number of individuals equal to $\sfrac{n}{k}$ -- which, as specified in Eqn.~(\ref{eq:n}), is $20$. This can be considered a good approximation over a statistically sufficient number of iterations of the algorithm. 

Let us analyze the complexity of each operator:
\begin{itemize}
    \item the bang operation expands $k$ centers of mass into a new population, ultimately generating a total of $n$ new individuals of size $d$: its complexity is $\mathcal{O}(nd)$;
    \item evaluating each individual in the population takes a linear time with the population size: its complexity is $\mathcal{O}(nd)$.
    \item practical implementations of k-means run in linear time with the dimensionality of the dataset and the number of clusters~\cite{pakhira2014linear} -- therefore, with a fixed number of iterations for each run, its complexity is $\mathcal{O}(knd)$; and
    \item the crunch operation simply retrieves the individual with minimum or maximum fitness value for each cluster by linearly scanning the entire population: its complexity is $\mathcal{O}(n)$.
\end{itemize}
With the relation reported in Eqn.~(\ref{eq:n}), we can replace any occurrence of $n$ with the term $k$ (big-o notation ignores constants; thus, we can remove the coefficient $20$). The overall complexity of a single run of k-BBBC is then expressed as in Eqn.~(\ref{eq:complexity}), in which k-means features the dominant term. Note that this analysis is valid for both the elitist and non-elitist algorithms, as the number of individuals in the population is increased only by a constant factor of $k$ and the elites do not undergo a second re-evaluation. 

\begin{equation}
    \label{eq:complexity}
    \begin{aligned}
    \mathcal{O}(kd+kd+k^2d+k) = \\\mathcal{O}(2kd+k+k^2d) = \\\mathcal{O}(k^2d)
    \end{aligned}
\end{equation}

The complexity can also be expressed in terms of the problem's properties, which include the parameters number of optima $m$ and dimensionality $d$. By applying the formulation in Eqn.~(\ref{eq:k}), we obtain a complexity as defined in Eqn.~(\ref{eq:complexity_problem}).

\begin{equation}
    \label{eq:complexity_problem}
    \begin{aligned}
    \mathcal{O}(k^2d) = \\\mathcal{O}((md)^2d) = \\\mathcal{O}(m^2d^3)
    \end{aligned}
\end{equation}

\section{Post Processing Methods}
\label{sec:postprocessing}

\subsection{Precise Identification of Detected Local Optima}
\label{sec:identification}

k-BBBC produces as output a population of individuals that will very likely contain more solutions than the number of local optima $m$. It will return exactly $k$ solutions (i.e., the centers of mass), which must overestimate the actual number of local optima for complete retrieval as specified in Sec.~\ref{sec:settings}. This will cause local optima as well as other points in their proximity to appear in the output. In other words, although the population of k-BBBC converges to the local optima, it is still necessary to identify them within the output.

As shown in Figure~\ref{fig:kbbbc_example_late}, k-BBBC converges to a well-separated population grouped around each local optima. Therefore, the simplest strategy to precisely identify the optimal solutions is using, once more, a clustering algorithm. 
The method described in Algorithm~\ref{alg:identification} proposes the following steps:
\begin{itemize}
    \item clustering the output with k-medoids having $k=m$ (line 2); and
    \item retrieving the optimal value within each cluster, which will be the $m$ local optimal solutions of the problem (lines 3-4, where \texttt{min} is used for minimization problems, and \texttt{max} for maximization).
\end{itemize} 

\IncMargin{1em}
\begin{algorithm}
        \SetKwData{P}{P}
        \SetKwData{O}{O}
        \SetKwData{opt}{m}
        \SetKwData{CM}{C}
        \SetKwFunction{KD}{kMedoids}
        \SetKwFunction{Min}{min}
        \SetKwFunction{Max}{max}
	\SetKwInOut{Input}{input}
	\SetKwInOut{Output}{output}
        \Input{Output of the optimization method $\P$, number of (expected) local optima of the problem $\opt$.}
	\Output{Array of local optima $\O$.}
	\BlankLine		
	
	\Begin{	
            $\P \leftarrow \KD(\P,\opt)$\;
            \ForEach{cluster \CM $\in$ \P}{%
                \If{minimization}{
                    add $\Min(\CM)$ in $\O$\;
                }
                \Else{
                    add $\Max(\CM)$ in $\O$\;
                }
            }
		\KwRet{$\O$};
	}
	
	\caption{Identification of Local Optima}\label{alg:identification}
\end{algorithm}\DecMargin{1em}

\begin{figure}[h]
    \centering
    {\includegraphics[width=\columnwidth]{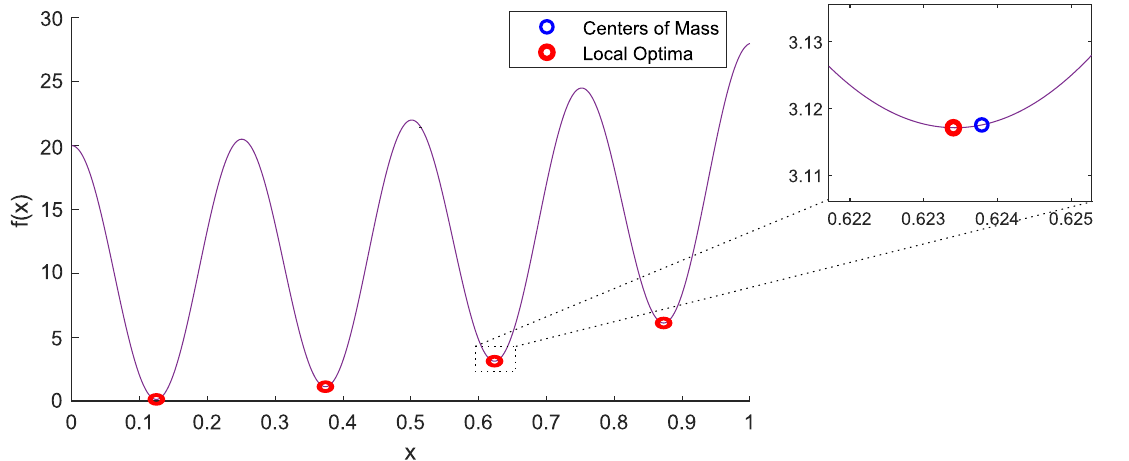}}\hfill
    \caption{Results of k-BBBC on a problem with $m=4$, showing the set of converged centers of mass in blue against the output of the identification method in red (i.e., the local optima). Since, on average, each concave region is divided between two clusters ($k=2md=8$), we can see a red point and a blue point for each region (the second region from the right is zoomed in to avoid graphical overlaps between points).}
    \label{fig:identification}
\end{figure}

For this procedure, we preferred k-medoids to k-means due to its lower error rate as specified in Sec.~\ref{sec:settings}. Figure~\ref{fig:identification} depicts the results of the identification method on the centers of mass retrieved by k-BBBC. 

\subsection{Quantifying the Number of Missed Optima}
\label{sec:quantification}

The procedure for optima identification described in Sec.~\ref{sec:identification} works well when the optimization method finds every local optimum in the search space. However, this is not always guaranteed: if the cardinality of the population $n$ is not large enough to explore the search space or if the number of clusters $k$ is insufficient, some local optima might be missed. When this happens, the $m$ identified solutions will contain repeated values as k-means is forced to divide the dataset into exactly $m$ clusters -- similarly to the issue described in Figure~\ref{fig:kmeans_error}, one or more clusters will be split among different solutions. This is purely dependent on the user-defined parameters chosen while optimizing a specific problem, as well as on the number of generations employed during the execution
 -- or sometimes, on the complexity of the problem that makes it difficult for some optima to be retrieved.

 \begin{figure}[h!]
    \centering
    {\includegraphics[width=8cm]{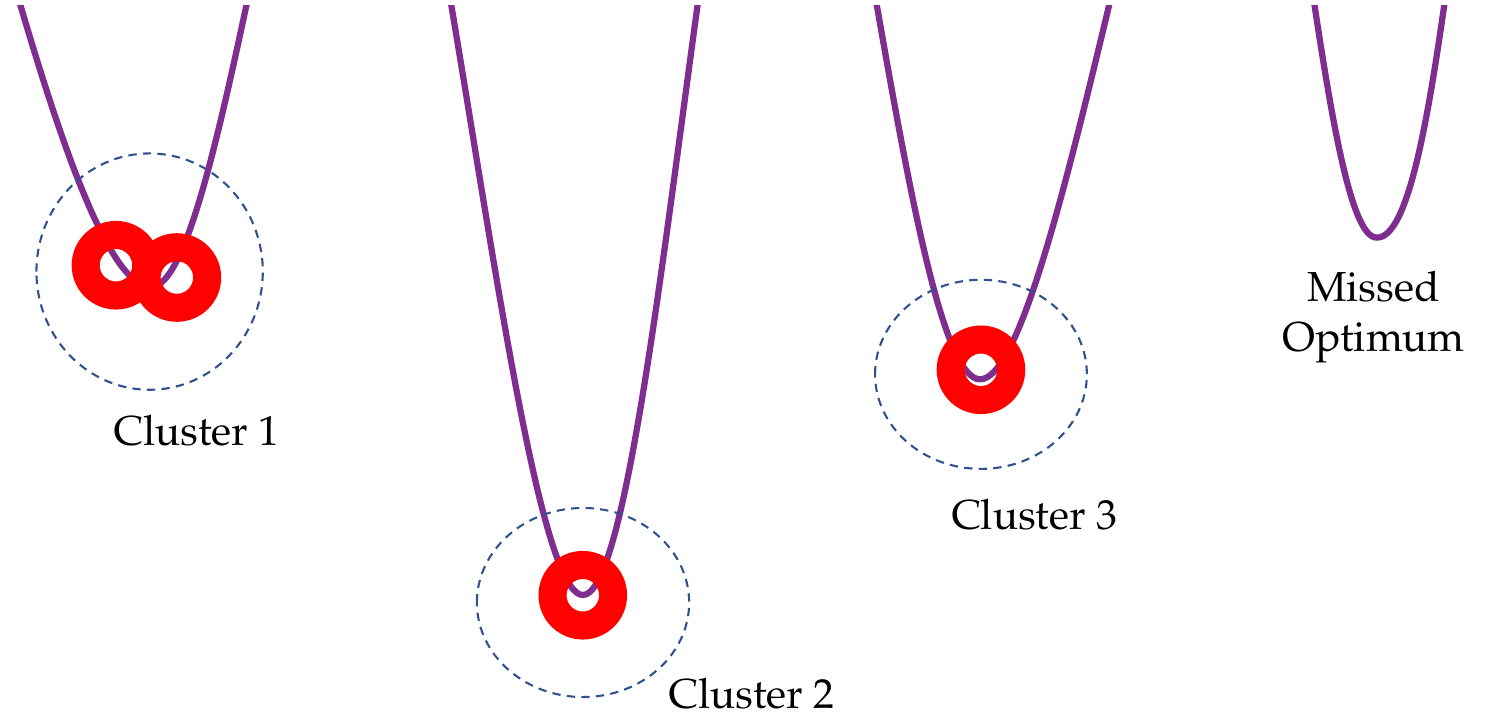}}\hfill
    \caption{If all $m$ optima of the problems are found, all solutions in the dataset converged to their respective optimum, and each of them identified its own cluster when divided into exactly $m$ clusters. However, in this example with four optima, the algorithm missed the rightmost optimum, and the identification method forced the leftmost optimum to be split into two points. As a result, the \textit{best} division of this dataset features three clusters instead of four.}
    \label{fig:missed_clusters}
\end{figure}

\begin{figure*}[p]
    \centering
    
    \subfigure[\protect\url{}\label{fig:detection_correct} Silhouette of complete detection.]%
    {\includegraphics[width=\linewidth]{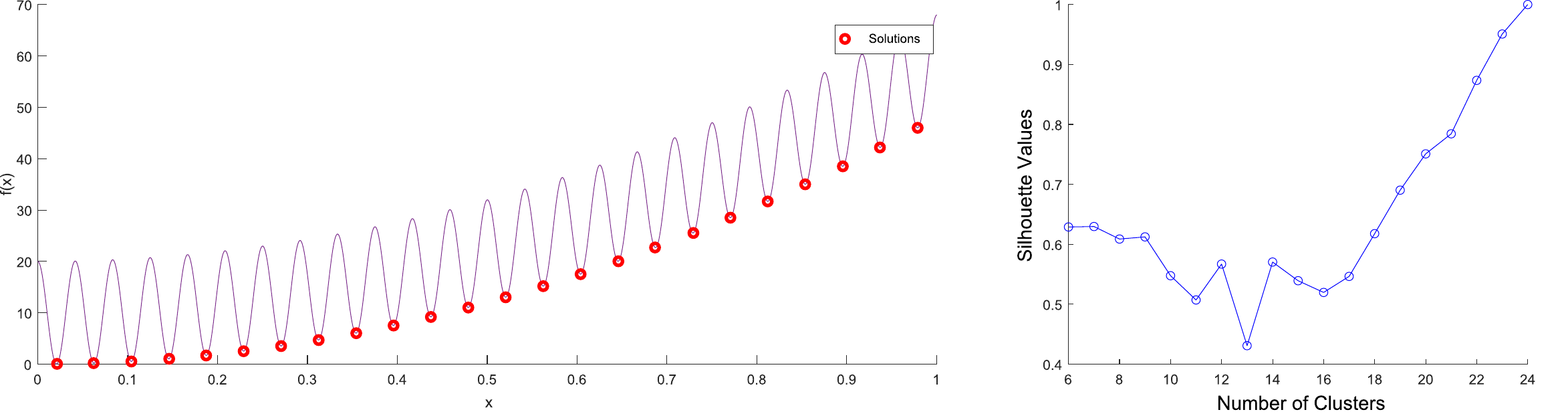}}
    
    \subfigure[\protect\url{}\label{fig:detection_fail_plateau} Silhouette of incomplete detection with plateau.]%
    {\includegraphics[width=\linewidth]{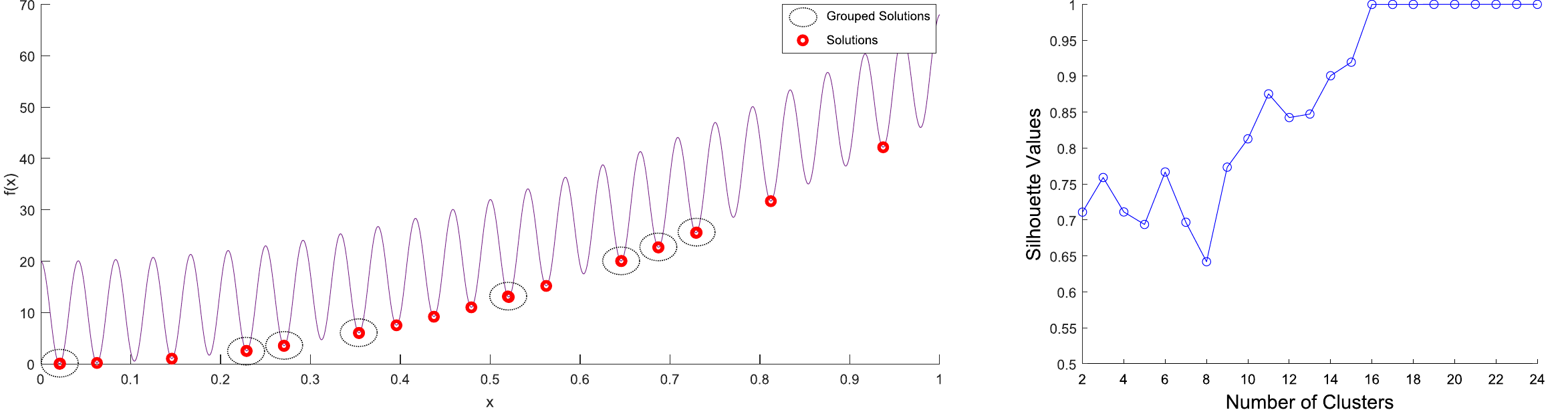}}
    
    \subfigure[\protect\url{}\label{fig:detection_fail_peak} Silhouette of incomplete detection with peak and  plateau.]%
    {\includegraphics[width=\linewidth]{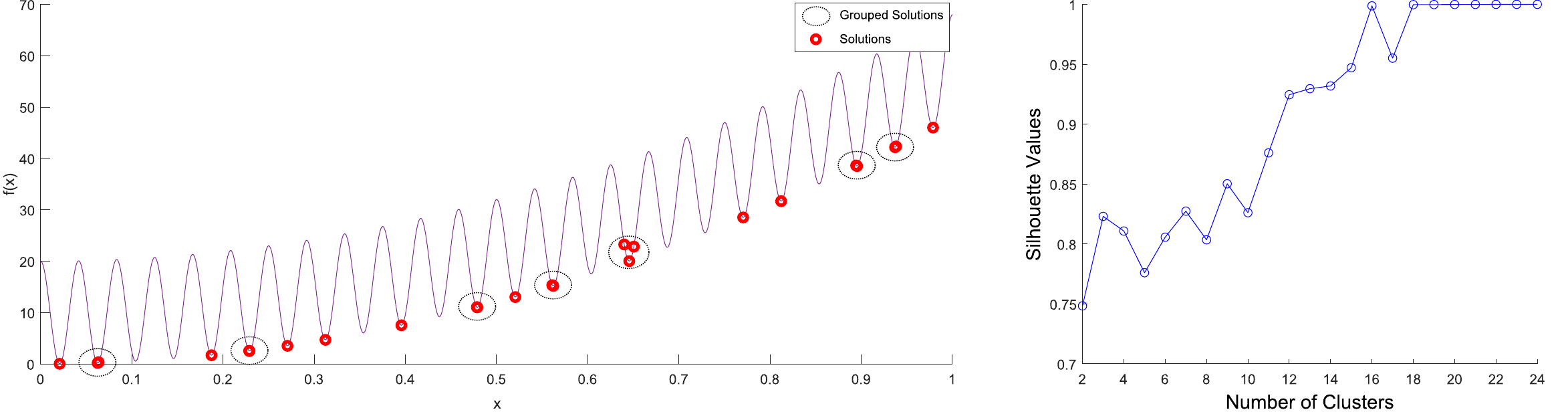}}
    
    \caption{The plots show the minimization of the Key function with $m=24$ along with its silhouette analysis. 
    In (a), k-BBBC was executed with $k=2m$ and the silhouette features a clear maximum value for $k=24$ -- no optima have been missed. 
    In (b), k-BBBC was executed with $k=m$, which was insufficient to locate all local optima. Therefore, some of the $k$ solutions retrieved by k-BBBC are very close to each other (like in Figure~\ref{fig:kmeans_error}). These solutions can be grouped in a single cluster; hence, the plateau in the silhouette ending at $k=16$ -- k-BBBC retrieved only $24-16=8$ local optima.   
    In (c), k-BBBC was executed with $k=m$ and ran for fewer generations than in (b), which was not sufficient to crunch the individuals to the local optima. Its silhouette features a plateau in the range $k=[18,24]$, and a peak on $k=16$ ($24-16=8$ missed optima): this is because the identification method (Sec.~\ref{sec:identification}) returned three points around the local optima at $0.66$, which can be divided into one cluster as well as into three clusters.
    }
    \label{fig:detection_examples}
\end{figure*}

Suppose we have a problem with $m=4$ optima, and the optimization algorithm missed one. Then, the identification method will return four solutions, two of which being very close to each other. When observing this dataset, one would divide it into $k=3$ clusters rather than $4$ due to the proximity of the two solutions identifying the same optimum. Of course, also $k=4=m$ would work, as each data in the dataset can be considered a cluster itself; however, the \textit{best number of clusters} would be $3$, indicating that the algorithm did miss one of them --  Figure~\ref{fig:missed_clusters} provides a graphical representation of this case. If the number of expected local optima $m$ is known, we then have a strategy to evaluate how many optima have been missed: observe the solutions retrieved by the identification method and determine the \textit{best} number of clusters $k^*$ we could divide them into. 
If $k^*<m$, the optimization algorithm missed $m-k^*$ optima.
Note that this process does not require knowing the optima's location in the search space and can be used for any problem. 

A way to estimate the \textit{best} number of clusters is using the silhouette score, which assesses the quality of clusters formed by k-means based on how effectively similar data points are grouped together~\cite{kaufman2009finding, rousseeuw1987silhouettes}. This criterion can find the correct value of $k^*$, by evaluating the output of k-means for each $k \in [2,m]$ -- the minimum number of clusters required for calculating silhouette score is $k=2$, because with $k=1$ we would retrieve one cluster containing the entire dataset. The silhouette score ranges from $-1$ to $1$; values of $k$ having a silhouette score near $1$ are considered a correct number of clusters for the given dataset. 
The silhouette score method can result in three cases -- we will indicate with $m_c$ the number of correctly identified optima, and with $m_m$ the number of missed optima:
\begin{enumerate}
    \item When the optimization algorithm correctly retrieves $m_c = m$ local optima, the silhouette values feature a maximum point at $k^* = m$. This is shown in Figure~\ref{fig:detection_correct}.
    \item When the optimization algorithm misses a total of $m_m$ local optima, the silhouette score for all the values of $k\in[m-m_m,m]$ will be close to $1$, as shown in Figure~\ref{fig:missed_clusters}. This results in a plateau of silhouette scores that represent different correct clusterizations of the given dataset; however, the value of $k^*$ is found at the left corner of the plateau. In Figure~\ref{fig:detection_fail_plateau}, a total of $m_m=8$ over $m=24$ optima were missed, and the silhouette trend features a plateau ending at $k^*=16$, resulting in $m-k^*=24-16=8=m_m$ missed optima and $k^*=16=m_c$ correct identified optima.
    \item When the optimization algorithm misses a total of $m_m$ local optima due to poor convergence, some values can be randomly scattered through the search space as shown in Figure~\ref{fig:detection_fail_peak}. In this case, the silhouette scores will feature the plateau preceded by a peak point, which is the value of $k^*$. However, this case is considered unstable and prone to detection mistakes: waiting for the optimization algorithm to converge correctly is recommended to avoid such occurrences. Note that the peak is visible only because the three solutions at $x = 0.6455$ are sufficiently close to each other to be considered as a cluster.
\end{enumerate}
This silhouette trend pertains to the $m$ solutions retrieved by the identification method for a given problem, which are datasets with a limited number of \textit{supposedly} well-separated points. Based on these observations, we developed the procedure illustrated in Algorithm~\ref{alg:missing_quantification} to detect the number of missed optima:

\IncMargin{1em}
\begin{algorithm}[b!]
        \SetKwData{E}{E}
        \SetKwData{O}{O}
        \SetKwData{opt}{m}
        \SetKwData{CM}{C}
        \SetKwData{S}{S}
        \SetKwFunction{Min}{min}
        \SetKwFunction{Max}{max}
        \SetKwFunction{eval}{evaluation}
        \SetKwFunction{slope}{slope}
        \SetKwFunction{Sil}{silhouette}
	\SetKwInOut{Input}{input}
	\SetKwInOut{Output}{output}
        \Input{Array of Detected Local optima $\O$ after the identification method, number of (expected) local optima of the problem $\opt$.}
	\Output{Number of correctly Detected Local optima $\opt_c$, number of missed local optima $\opt_m$.}
	\BlankLine		
	
            

        \Begin{	
            $\E \leftarrow \Sil(\O,2:\opt)$\;
            $\opt_c \leftarrow \opt$\;
            \For{$k \in [\opt-1, 2]$}
            {
                \If{$\left(\E(\opt) - \E(k) \right) / \left(1 - \frac{k-1}{\opt-1} \right) < 0.1$}{
                    $\opt_c \leftarrow k$\;
                }
            }
            $\opt_m \leftarrow \opt - \opt_c;$
            
		\KwRet{$\opt_c, \opt_m $};
	}
	
	\caption{Quantification of Missed Optima}\label{alg:missing_quantification}
\end{algorithm}\DecMargin{1em}

\begin{itemize}
    \item evaluate the silhouette scores for each value of $k \in [2, m]$ (line 2);
    \item start the procedure assuming that the algorithm correctly retrieved all optima (line 3);
    \item we evaluate the slope of the segments having as edges the silhouette score at $k=m$ (the last one) and, iteratively, each $k$ from the rightmost to the leftmost (lines 4 and 5) -- this allows us to consider as a straight line (i.e., a plateau) also multiple consecutive scores in a row, as well as being able to detect any high peak (by treating them as a continuation of the plateau);
    \item if the slope of the segment for a specific $k$ is smaller than a threshold (empirically set to $0.1$), then we can consider the silhouette trend as a straight line so far (i.e., we missed some optima), and we change the value of $m_c$ to the current $k$ (lines 5 and 6);
    \item if the slope is bigger than the threshold, then we are not in the plateau and the value of $m_c$ is not updated; and
    \item at the end of the loop, we evaluate the number of missed optima by subtracting the value of $m_c$ (i.e., $k^*$) from the total number of optima of the problem $m$ (line 7).
\end{itemize}
The formula for slope evaluation at line 5 maintains the ratio between silhouette scores and the number of clusters $k$ to $1:1$, considering the range of silhouette scores to be equal to 1 (even though the state-of-the-art reports that the lower bound is $-1$, in all our trials we have never observed negative scores). This helped define the correct threshold for the plateau. 

Note that, for this detection method to work, the population of the optimization algorithm must have converged to the local optima. 

\section{Experiments for Finding Local Peaks (on Basic Multimodal Problems)}
\label{sec:experiments_local}

In this section, we report the results of k-BBBC (both elitist and not) compared to another method in the literature: Multimodal Cuckoo Search (MCS)~\cite{cuevas2014cuckoo}. We selected only this algorithm for two reasons: (i) based on the original work, it outperforms other multimodal algorithms such as CDE~\cite{thomsen2004multimodal}, FSDE~\cite{miller1996genetic,thomsen2004multimodal}, CP~\cite{petrowski1996clearing}, EPS~\cite{liang2011genetic}, CSA~\cite{de2002learning}, and AIN~\cite{de2002artificial} -- so we did not need to implement each of them for testing; and (ii) its code is publicly available on MathWorks File Exchange.

For k-BBBC, we ran the post-processing described in Sec.~\ref{sec:postprocessing} -- that is because k-BBBC ensures that the entire population converges to the optima. However, for MCS we ran a different post-processing, since it does not ensure the same trend: after termination, we executed its \textit{depuration} procedure~\cite{cuevas2014cuckoo} to identify the local optima within the retrieved population -- similarly to our identification procedure described in Sec.~\ref{sec:identification}. This should return a number of individuals that is $<=m$; however, in case this number is larger, we executed our identification procedure. Lastly, this dataset of individuals goes through the quantification method of Sec.~\ref{sec:quantification} to assess how many optima have been missed.

\subsection{Methodology}
\label{sec:experiment_methodology}

The algorithms were compared on the twenty low-dimensionality (i.e., $d=1$ or $2$) and four high-dimensionality problems (i.e., $d>4$).
We executed every algorithm with the same population size, defined as in Eqn.~(\ref{eq:k}) and (\ref{eq:n}) and based on the problem to be optimized. To ensure a fair comparison, we executed k-BBBC for $1000$ generations, recorded its runtime for each function, and set it as a stopping condition for MCS (i.e., we stopped MCS at the end of the iteration that exceeded k-BBBC runtime -- specifically, the average runtime of non-elitist k-BBBC). Although the authors did not report it, we observed that MCS has a higher time complexity than k-BBBC due to its \textit{depuration} procedure; therefore, setting a fixed number of population and generations would have resulted in an unfeasible runtime for problems with a large number of optima.
Every problem was solved $25$ times per algorithm, and every result is presented as the average and standard deviation of these runs.

We conducted all experiments on a computer with a 4-core 2.80 GHz CPU and 16 GB RAM.

\subsection{Evaluation Metrics}
\label{sec:experiments_local_methodology}

We evaluated the algorithms' accuracy in both search (\ref{eq:accuracy_search}) and objective space (\ref{eq:accuracy_objective}) by comparing each retrieved optima $\textbf{x}$ with the exact ones $\textbf{z}$. 

\begin{equation}
    \label{eq:accuracy_search}
    \begin{aligned}
        a_{src} = \sum_{i=1}^{m}{\norm{\textbf{x}_i - \textbf{z}_i}^2}
    \end{aligned}
\end{equation}

\begin{equation}
    \label{eq:accuracy_objective}
    \begin{aligned}
        a_{obj} = \sum_{i=1}^{m}{|f(\textbf{x}_i) - f(\textbf{z}_i)|}
    \end{aligned}
\end{equation}

Additionally, we report the number of detected optima $\bar{m}$ over the expected value $m$. We considered an optimum $\textbf{z}$ to be correctly detected when we find a solution $\textbf{x}$ within a specific radius from it -- in the search space. This radius was manually defined for each function by measuring the base of its thinner concave region.

We also validated the quantification method described in Algorithm~\ref{alg:missing_quantification} by comparing the success rate of its output $\sfrac{m_c}{m}$ with the one from the actual counting $\sfrac{\bar{m}}{m}$ -- which is considered to be correct. If these two values are the same, then the quantification method is a valid algorithm to use when solving a problem whose optima are unknown. 

Lastly, we report the runtime in seconds (which is set to be approximately the same for all algorithms) and the number of function evaluations during execution (i.e., how many times the algorithm calculates the value of the objective function).

\subsection{Experiments with low dimensionality problems}
\label{sec:low_dim}

\begin{table*}[!t]
\centering
\caption{Basic Low-Dimension Multimodal Problems}
\label{tab:bench_functions}
\resizebox{\textwidth}{!}{%
\begin{tabular}{|ccccccc|}
\hline
\textbf{Function} & \textbf{Type} & \textbf{Bounds} & \makecell{\textbf{Dimensions}\\$(d)$} & \makecell{\textbf{Optima}\\$(m)$} & \makecell{\textbf{Clusters}\\($k$)} & \makecell{\textbf{Population Size}\\($n$)}\\ \hline

\begin{tabular}{l}
\makecell{Uneven Decreasing Maxima \\ $f_1(\textbf{x}) = 2^{-2(\frac{x-0.1}{0.8})^2}\cdot \sin{5 \pi x}^6$}
\end{tabular}
& 
\begin{tabular}{l}
\texttt{max} 
\end{tabular}
& 
\begin{tabular}{l}
$\left[0,1\right]$ 
\end{tabular}
& 
\begin{tabular}{l}
$1$ 
\end{tabular}
&
\begin{tabular}{l}
$5$ 
\end{tabular}
&
\begin{tabular}{l}
$10$ 
\end{tabular}
&
\begin{tabular}{l}
$200$ 
\end{tabular}
\\ \hline

\begin{tabular}{l}
\makecell{Key4 \\ $f_2(\textbf{x}) = {10 (1+\cos{(2 \pi 4 x)}) + 8 x^2}$}
\end{tabular}
& 
\begin{tabular}{l}
\texttt{min} 
\end{tabular}
& 
\begin{tabular}{l}
$\left[0,1\right]$ 
\end{tabular}
& 
\begin{tabular}{l}
$1$ 
\end{tabular}
&
\begin{tabular}{l}
$4$ 
\end{tabular}
&
\begin{tabular}{l}
$8$ 
\end{tabular}
&
\begin{tabular}{l}
$160$ 
\end{tabular}
\\ \hline

\begin{tabular}{l}
\makecell{Key8 \\ $f_3(\textbf{x}) = {10 (1+\cos{(2 \pi 8 x)}) + 16 x^2}$}
\end{tabular}
& 
\begin{tabular}{l}
\texttt{min} 
\end{tabular}
& 
\begin{tabular}{l}
$\left[0,1\right]$ 
\end{tabular}
& 
\begin{tabular}{l}
$1$ 
\end{tabular}
&
\begin{tabular}{l}
$8$ 
\end{tabular}
&
\begin{tabular}{l}
$16$ 
\end{tabular}
&
\begin{tabular}{l}
$320$ 
\end{tabular}
\\ \hline

\begin{tabular}{l}
\makecell{Key16 \\ $f_4(\textbf{x}) = {10 (1+\cos{(2 \pi 16 x)}) + 32 x^2}$}
\end{tabular}
& 
\begin{tabular}{l}
\texttt{min} 
\end{tabular}
& 
\begin{tabular}{l}
$\left[0,1\right]$ 
\end{tabular}
& 
\begin{tabular}{l}
$1$ 
\end{tabular}
&
\begin{tabular}{l}
$16$ 
\end{tabular}
&
\begin{tabular}{l}
$32$ 
\end{tabular}
&
\begin{tabular}{l}
$640$ 
\end{tabular}
\\ \hline

\begin{tabular}{l}
\makecell{Key24 \\ $f_5(\textbf{x}) = {10 (1+\cos{(2 \pi 24 x)}) + 48 x^2}$}
\end{tabular}
& 
\begin{tabular}{l}
\texttt{min} 
\end{tabular}
& 
\begin{tabular}{l}
$\left[0,1\right]$ 
\end{tabular}
& 
\begin{tabular}{l}
$1$ 
\end{tabular}
&
\begin{tabular}{l}
$24$ 
\end{tabular}
&
\begin{tabular}{l}
$48$ 
\end{tabular}
&
\begin{tabular}{l}
$960$ 
\end{tabular}
\\ \hline

\begin{tabular}{l}
\makecell{Key48 \\ $f_6(\textbf{x}) = {10 (1+\cos{(2 \pi 48 x)}) + 96 x^2}$}
\end{tabular}
& 
\begin{tabular}{l}
\texttt{min} 
\end{tabular}
& 
\begin{tabular}{l}
$\left[0,1\right]$ 
\end{tabular}
& 
\begin{tabular}{l}
$1$ 
\end{tabular}
&
\begin{tabular}{l}
$48$ 
\end{tabular}
&
\begin{tabular}{l}
$96$ 
\end{tabular}
&
\begin{tabular}{l}
$1920$ 
\end{tabular}
\\ \hline

\begin{tabular}{l}
\makecell{Key96 \\ $f_7(\textbf{x}) = {10 (1+\cos{(2 \pi 96 x)}) + 192 x^2}$}
\end{tabular}
& 
\begin{tabular}{l}
\texttt{min} 
\end{tabular}
& 
\begin{tabular}{l}
$\left[0,1\right]$ 
\end{tabular}
& 
\begin{tabular}{l}
$1$ 
\end{tabular}
&
\begin{tabular}{l}
$96$ 
\end{tabular}
&
\begin{tabular}{l}
$192$ 
\end{tabular}
&
\begin{tabular}{l}
$3840$ 
\end{tabular}
\\ \hline

\begin{tabular}{l}
\makecell{Schwefel \\ $f_8(\textbf{x}) = 418.9829 - x \sin{(\sqrt{|x|})}$}
\end{tabular}
& 
\begin{tabular}{l}
\texttt{min} 
\end{tabular}
& 
\begin{tabular}{l}
$\left[-500,500\right]$ 
\end{tabular}
& 
\begin{tabular}{l}
$1$ 
\end{tabular}
&
\begin{tabular}{l}
$8$ 
\end{tabular}
&
\begin{tabular}{l}
$16$ 
\end{tabular}
&
\begin{tabular}{l}
$320$ 
\end{tabular}
\\ \hline

\begin{tabular}{l}
\makecell{Schwefel \\ $f_9(\textbf{x}) = 2 \cdot 418.9829 - \sum_{i=1}^{2}{x_i\sin{(\sqrt{|x_i|})}}$}
\end{tabular}
& 
\begin{tabular}{l}
\texttt{min} 
\end{tabular}
& 
\begin{tabular}{l}
$\left[-500,500\right]$ 
\end{tabular}
& 
\begin{tabular}{l}
$2$ 
\end{tabular}
&
\begin{tabular}{l}
$64$ 
\end{tabular}
&
\begin{tabular}{l}
$256$ 
\end{tabular}
&
\begin{tabular}{l}
$5120$ 
\end{tabular}
\\ \hline

\begin{tabular}{l}
\makecell{Himmelblau \\ $f_{10}(\textbf{x}) = {({x_1}^2 + x_2 - 11)^2 +(x_1 + {x_2}^2 - 7)^2}$}
\end{tabular}
& 
\begin{tabular}{l}
\texttt{min} 
\end{tabular}
& 
\begin{tabular}{l}
$\left[-6,6\right]$ 
\end{tabular}
& 
\begin{tabular}{l}
$2$ 
\end{tabular}
&
\begin{tabular}{l}
$4$ 
\end{tabular}
&
\begin{tabular}{l}
$16$ 
\end{tabular}
&
\begin{tabular}{l}
$320$ 
\end{tabular}
\\ \hline

\begin{tabular}{l}
\makecell{Bird \\ $f_{11}(\textbf{x}) = {\sin(x_1)\exp(1 - \cos{x_2})^2 +\cos{x_2}\exp(1 - \sin{x_1})^2 + (x_1 - x_2)^2}$}
\end{tabular}
& 
\begin{tabular}{l}
\texttt{min} 
\end{tabular}
& 
\begin{tabular}{l}
$\left[-2\pi,2\pi\right]$ 
\end{tabular}
& 
\begin{tabular}{l}
$2$ 
\end{tabular}
&
\begin{tabular}{l}
$4$ 
\end{tabular}
&
\begin{tabular}{l}
$16$ 
\end{tabular}
&
\begin{tabular}{l}
$320$ 
\end{tabular}
\\ \hline

\begin{tabular}{l}
\makecell{Rastrigin \\ $f_{12}(\textbf{x}) = {20 + {x_1}^2 + {x_2}^2 } -  10(\cos{(2\pi x_1)} + \cos{(2\pi x_2)})$}
\end{tabular}
& 
\begin{tabular}{l}
\texttt{min} 
\end{tabular}
& 
\begin{tabular}{l}
$\left[-5.12,5.12\right]$ 
\end{tabular}
& 
\begin{tabular}{l}
$2$ 
\end{tabular}
&
\begin{tabular}{l}
$121$ 
\end{tabular}
&
\begin{tabular}{l}
$484$ 
\end{tabular}
&
\begin{tabular}{l}
$9680$ 
\end{tabular}
\\ \hline

\begin{tabular}{l}
\makecell{Cosine Mixture \\ $f_{13}(\textbf{x}) =  {-(0.2 + {x_1}^2 + {x_2}^2 - 0.1\cos{(6\pi x_1)} - 0.1. \cos{(6 \pi x_2)})}$}
\end{tabular}
& 
\begin{tabular}{l}
\texttt{min} 
\end{tabular}
& 
\begin{tabular}{l}
$\left[-1,1\right]$ 
\end{tabular}
& 
\begin{tabular}{l}
$2$ 
\end{tabular}
&
\begin{tabular}{l}
$25$ 
\end{tabular}
&
\begin{tabular}{l}
$100$ 
\end{tabular}
&
\begin{tabular}{l}
$2000$ 
\end{tabular}
\\ \hline

\begin{tabular}{l}
\makecell{Cross-in-Tray \\ $f_{14}(\textbf{x}) = {-(10^{-4}\abs{\sin{x_1}\sin{x_2}\exp(\abs{100 - \frac{\sqrt{{x_1}^2 + {x_2}^2}}{\pi}})} + 1)^{0.1}}$}
\end{tabular}
& 
\begin{tabular}{l}
\texttt{min} 
\end{tabular}
& 
\begin{tabular}{l}
$\left[-9.5,9.5\right]$ 
\end{tabular}
& 
\begin{tabular}{l}
$2$ 
\end{tabular}
&
\begin{tabular}{l}
$36$ 
\end{tabular}
&
\begin{tabular}{l}
$144$ 
\end{tabular}
&
\begin{tabular}{l}
$2880$ 
\end{tabular}
\\ \hline

\begin{tabular}{l}
\makecell{Vincent \\ $f_{15}(\textbf{x}) = {-(\sin{(10\log{x_1})} + \sin{(10\log{x_2})})}$}
\end{tabular}
& 
\begin{tabular}{l}
\texttt{min} 
\end{tabular}
& 
\begin{tabular}{l}
$\left[0.25,10\right]$ 
\end{tabular}
& 
\begin{tabular}{l}
$2$ 
\end{tabular}
&
\begin{tabular}{l}
$36$ 
\end{tabular}
&
\begin{tabular}{l}
$144$ 
\end{tabular}
&
\begin{tabular}{l}
$2880$ 
\end{tabular}
\\ \hline

\begin{tabular}{l}
\makecell{Holder Table \\ $f_{16}(\textbf{x}) = {-\abs{\sin{x_1}\cos{x_2}\exp(\abs{1 - \frac{\sqrt{{x_1}^2 + {x_2}^2}}{\pi}})}}$}
\end{tabular}
& 
\begin{tabular}{l}
\texttt{min} 
\end{tabular}
& 
\begin{tabular}{l}
$\left[-10,10\right]$ 
\end{tabular}
& 
\begin{tabular}{l}
$2$ 
\end{tabular}
&
\begin{tabular}{l}
$56$ 
\end{tabular}
&
\begin{tabular}{l}
$224$ 
\end{tabular}
&
\begin{tabular}{l}
$4480$ 
\end{tabular}
\\ \hline

\begin{tabular}{l}
\makecell{Pen Holder \\ $f_{17}(\textbf{x}) = {-\exp(-|\cos{x_1}\cos{x_2}\cdot \exp(\abs{1 - \frac{\sqrt{{x_1}^2 + {x_2}^2}}{\pi}})|^{-1})}$}
\end{tabular}
& 
\begin{tabular}{l}
\texttt{max} 
\end{tabular}
& 
\begin{tabular}{l}
$\left[0,1\right]$ 
\end{tabular}
& 
\begin{tabular}{l}
$2$ 
\end{tabular}
&
\begin{tabular}{l}
$49$ 
\end{tabular}
&
\begin{tabular}{l}
$196$ 
\end{tabular}
&
\begin{tabular}{l}
$3920$ 
\end{tabular}
\\ \hline

\begin{tabular}{l}
\makecell{Egg Crate \\ $f_{18}(\textbf{x}) = {{x_1}^2 + {x_2}^2 + 25(\sin{x_1}^2 + \sin{x_2}^2)}$}
\end{tabular}
& 
\begin{tabular}{l}
\texttt{min} 
\end{tabular}
& 
\begin{tabular}{l}
$\left[-5,5\right]$ 
\end{tabular}
& 
\begin{tabular}{l}
$2$ 
\end{tabular}
&
\begin{tabular}{l}
$9$ 
\end{tabular}
&
\begin{tabular}{l}
$36$ 
\end{tabular}
&
\begin{tabular}{l}
$720$ 
\end{tabular}
\\ \hline

\begin{tabular}{l}
\makecell{Griewank \\ $f_{20}(\textbf{x}) = {\frac{{x_1}^2 + {x_2}^2}{4000} - \cos{x}\cos{(\frac{x_2}{\sqrt{2}})} + 1}$}
\end{tabular}
& 
\begin{tabular}{l}
\texttt{min} 
\end{tabular}
& 
\begin{tabular}{l}
$\left[-50,50\right]$ 
\end{tabular}
& 
\begin{tabular}{l}
$2$ 
\end{tabular}
&
\begin{tabular}{l}
$379$ 
\end{tabular}
&
\begin{tabular}{l}
$1516$ 
\end{tabular}
&
\begin{tabular}{l}
$30320$ 
\end{tabular}
\\ \hline

\begin{tabular}{l}
\makecell{Griewank \\ $f_{20}(\textbf{x}) = {-\frac{{x_1}^2 + {x_2}^2}{4000} + \cos{x}\cos{(\frac{x_2}{\sqrt{2}})} - 1}$}
\end{tabular}
& 
\begin{tabular}{l}
\texttt{max} 
\end{tabular}
& 
\begin{tabular}{l}
$\left[-50,50\right]$ 
\end{tabular}
& 
\begin{tabular}{l}
$2$ 
\end{tabular}
&
\begin{tabular}{l}
$379$ 
\end{tabular}
&
\begin{tabular}{l}
$1516$ 
\end{tabular}
&
\begin{tabular}{l}
$30320$ 
\end{tabular}
\\ \hline

\end{tabular}%
}
\end{table*}

The twenty problems with low dimensionality are summarized in Table~\ref{tab:bench_functions}. For each problem, the exact local optima $\textbf{z}$ were manually retrieved by running \texttt{fmincon} (Interior Point Algorithm~\cite{byrd1999interior}) on the search space's sub-regions surrounding them -- i.e., we manually located each concave region by observing the function's graph and set its geometrical limits as bounds of the problem. 

\begin{table*}[h!]
\captionsetup{justification=centering}
\caption{Results for Low Dimensionality Problems}
\label{tab:results_low}
\resizebox{\linewidth}{!}{%
\begin{tabular}{c|c|c|c|c|c|c|c|c|}
\cline{2-9}
\multicolumn{1}{l|}{}                                   & \makecell{\textbf{Algorithm}} & \makecell{\textbf{Accuracy} \\ \textbf{Search Space} \\ $a_{src}$}           & \makecell{\textbf{Accuracy} \\ \textbf{Objective Space} \\ $a_{obj}$}   & \makecell{\textbf{Detected} \\ \textbf{Local Optima} \\ $\bar{m}$}    & \makecell{ \textbf{Success Rate} \\ Actual \\ $\sfrac{\bar{m}}{m}$ } & \makecell{ \textbf{Success Rate} \\ Quantification \\ $\sfrac{m_c}{m}$}  & \makecell{\textbf{Runtime} \\ $\left[\text{seconds}\right]$}   & \makecell{\textbf{Num. of Function} \\ \textbf{Evaluations}}
 \\ \hline

\multicolumn{1}{|c|}{\multirow{3}{*}{\rotatebox[origin=c]{90}{\textbf{$f_1$\color[HTML]{FFFFFF}}}}} & 
k-BBBC   & $\expnumber{1.32}{-4} \pm \expnumber{5.63}{-5}$ & $\expnumber{3.52}{-6} \pm \expnumber{3.86}{-6}$ & $5.00 \pm 0.00$ & $1.00 \pm 0.00$ & $1.00 \pm 0.00$ & $3.39\pm 0.15$ & $200000 \pm 0.00$ \\
\multicolumn{1}{|c|}{} & 
E-k-BBBC  & $\expnumber{2.51}{-7} \pm \expnumber{1.02}{-7}$ & $\expnumber{5.98}{-10} \pm \expnumber{7.66}{-12}$ & $5.00 \pm 0.00$ & $1.00 \pm 0.00$ & $1.00 \pm 0.00$ & $3.10\pm 0.08$ & \ $200000 \pm 0.00$ \\
\multicolumn{1}{|c|}{} &  
MCS   & $\expnumber{1.63}{-6} \pm \expnumber{1.61}{-6}$ & $\expnumber{2.04}{-9} \pm \expnumber{3.13}{-9}$ & $4.00 \pm 0.00$ & $0.80\pm 0.00$ & $0.80\pm 0.00$ & $3.39\pm 0.01$ & $218799.20 \pm 10168.27$ \\
\hline 
\hline

\multicolumn{1}{|c|}{\multirow{3}{*}{\rotatebox[origin=c]{90}{\textbf{$f_2$\color[HTML]{FFFFFF}}}}} & 
k-BBBC   & $\expnumber{1.19}{-4} \pm \expnumber{6.31}{-5}$ & $\expnumber{2.52}{-5} \pm \expnumber{3.48}{-5}$ & $4.00 \pm 0.00$ & $1.00 \pm 0.00$ & $1.00 \pm 0.00$ & $2.96\pm 0.07$ & $160000 \pm 0.00$ \\
\multicolumn{1}{|c|}{} & 
E-k-BBBC   & $\expnumber{1.75}{-7} \pm \expnumber{8.35}{-8}$ & $\expnumber{1.10}{-9} \pm \expnumber{6.15}{-11}$ & $4.00 \pm 0.00$ & $1.00 \pm 0.00$ & $1.00 \pm 0.00$ & $2.71\pm 0.07$ & $160000 \pm 0.00$ \\
\multicolumn{1}{|c|}{} &  
MCS   & $\expnumber{1.78}{-6} \pm \expnumber{1.15}{-6}$ & $\expnumber{9.14}{-9} \pm \expnumber{1.17}{-8}$ & $4.00 \pm 0.00$ & $1.00 \pm 0.00$ & $1.00 \pm 0.00$ & $2.97\pm 0.01$ & $155014.88 \pm 6893.55$ \\
\hline 
\hline

\multicolumn{1}{|c|}{\multirow{3}{*}{\rotatebox[origin=c]{90}{\textbf{$f_3$\color[HTML]{FFFFFF}}}}} & 
k-BBBC   & $\expnumber{2.45}{-4} \pm \expnumber{1.10}{-4}$ & $\expnumber{2.62}{-4} \pm \expnumber{3.86}{-4}$ & $8.00 \pm 0.00$ & $1.00 \pm 0.00$ & $1.00 \pm 0.00$ & $4.33\pm 0.09$ & $320000 \pm 0.00$ \\
\multicolumn{1}{|c|}{} & 
E-k-BBBC   & $\expnumber{3.92}{-7} \pm \expnumber{1.33}{-7}$ & $\expnumber{3.52}{-9} \pm \expnumber{2.93}{-10}$ & $8.00 \pm 0.00$ & $1.00 \pm 0.00$ & $1.00 \pm 0.00$ & $4.21\pm 0.14$ & $320000 \pm 0.00$ \\
\multicolumn{1}{|c|}{} &  
MCS   & $\expnumber{8.43}{-6} \pm \expnumber{4.90}{-6}$ & $\expnumber{4.85}{-7} \pm \expnumber{7.37}{-7}$ & $8.00 \pm 0.00$ & $1.00 \pm 0.00$ & $1.00 \pm 0.00$ & $4.34\pm 0.01$ & $245645.88 \pm 9737.08$ \\
\hline 
\hline

\multicolumn{1}{|c|}{\multirow{3}{*}{\rotatebox[origin=c]{90}{\textbf{$f_4$\color[HTML]{FFFFFF}}}}} & 
k-BBBC   & $\expnumber{4.74}{-4} \pm \expnumber{1.39}{-4}$ & $\expnumber{1.76}{-3} \pm \expnumber{1.35}{-3}$ & $15.96  \pm 0.20$ & $0.99 \pm 0.01$ & $0.99 \pm 0.01$ & $7.25\pm 0.18$ & $640000 \pm 0.00$ \\
\multicolumn{1}{|c|}{} & 
E-k-BBBC   & $\expnumber{8.44}{-7} \pm \expnumber{2.59}{-7}$ & $\expnumber{2.58}{-8} \pm \expnumber{2.34}{-9}$ & $16.00 \pm 0.00$ & $1.00 \pm 0.00$ & $1.00 \pm 0.00$ & $7.40\pm 0.16$ & $640000 \pm 0.00$ \\
\multicolumn{1}{|c|}{} &  
MCS   & $\expnumber{2.68}{-5} \pm \expnumber{1.12}{-5}$ & $\expnumber{1.29}{-5} \pm \expnumber{1.74}{-5}$ & $16.00 \pm 0.00$ & $1.00 \pm 0.00$ & $1.00 \pm 0.00$ & $7.26\pm 0.01$ & $392039.48 \pm 13542.77$ \\
\hline 
\hline

\multicolumn{1}{|c|}{\multirow{3}{*}{\rotatebox[origin=c]{90}{\textbf{$f_5$\color[HTML]{FFFFFF}}}}} & 
k-BBBC   & $\expnumber{7.15}{-4} \pm \expnumber{1.67}{-4}$ & $\expnumber{5.47}{-3} \pm \expnumber{2.74}{-3}$ & $24.00  \pm 0.00$ & $1.00 \pm 0.00$ & $1.00 \pm 0.00$ & $10.93\pm 0.25$ & $960000 \pm 0.00$ \\
\multicolumn{1}{|c|}{} & 
E-k-BBBC   & $\expnumber{1.30}{-6} \pm \expnumber{3.17}{-7}$ & $\expnumber{5.01}{-8} \pm \expnumber{6.84}{-9}$ & $24.00  \pm 0.00$ & $1.00 \pm 0.00$ & $1.00 \pm 0.00$ & $11.13\pm 0.17$ & $960000 \pm 0.00$ \\
\multicolumn{1}{|c|}{} &  
MCS   & $\expnumber{5.12}{-5} \pm \expnumber{6.37}{-5}$ & $\expnumber{5.90}{-4} \pm \expnumber{1.88}{-3}$ & $23.08  \pm 0.39$ & $0.96 \pm 0.01$ & $0.96 \pm 0.01$ & $10.96\pm 0.02$ & $528258.12 \pm 19685.27$ \\
\hline 
\hline

\multicolumn{1}{|c|}{\multirow{3}{*}{\rotatebox[origin=c]{90}{\textbf{$f_6$\color[HTML]{FFFFFF}}}}} & 
k-BBBC   & $\expnumber{1.34}{-3} \pm \expnumber{1.93}{-4}$ & $\expnumber{3.83}{-2} \pm \expnumber{1.60}{-2}$ & $48.00  \pm 0.00$ & $1.00 \pm 0.00$ & $1.00 \pm 0.00$ & $27.25\pm 0.57$ & $1920000 \pm 0.00$ \\
\multicolumn{1}{|c|}{} & 
E-k-BBBC   & $\expnumber{2.46}{-6} \pm \expnumber{3.17}{-7}$ & $\expnumber{1.09}{-7} \pm \expnumber{3.16}{-8}$ & $48.00  \pm 0.00$ & $1.00 \pm 0.00$ & $1.00 \pm 0.00$ & $26.30\pm 0.52$ & $1920000 \pm 0.00$ \\
\multicolumn{1}{|c|}{} &  
MCS   & $\expnumber{1.26}{-1} \pm \expnumber{2.29}{-3}$ & $\expnumber{2.32}{+2} \pm \expnumber{1.61}{+1}$ & $21.64  \pm 0.48$ & $0.45 \pm 0.01$ & $0.89 \pm 0.01$ & $27.66\pm 0.04$ & $853305.52 \pm 104474.40$ \\
\hline 
\hline

\multicolumn{1}{|c|}{\multirow{3}{*}{\rotatebox[origin=c]{90}{\textbf{$f_7$\color[HTML]{FFFFFF}}}}} & 
k-BBBC   & $\expnumber{3.81}{-3} \pm \expnumber{2.54}{-4}$ & $\expnumber{3.27}{-1} \pm \expnumber{7.25}{-2}$ & $96.00  \pm 0.00$ & $1.00 \pm 0.00$ & $1.00 \pm 0.00$ & $76.97\pm 1.80$ & $3840000 \pm 0.00$ \\
\multicolumn{1}{|c|}{} & 
E-k-BBBC   & $\expnumber{2.41}{-3} \pm \expnumber{7.11}{-7}$ & $\expnumber{2.48}{-3} \pm \expnumber{2.32}{-7}$ & $96.00  \pm 0.00$ & $1.00 \pm 0.00$ & $1.00 \pm 0.00$ & $74.97\pm 2.22$ & $3840000 \pm 0.00$ \\
\multicolumn{1}{|c|}{} &  
MCS   & $\expnumber{2.08}{-3} \pm \expnumber{1.71}{-5}$ & $\expnumber{2.41}{-3} \pm \expnumber{7.09}{-4}$ & $82.36  \pm 0.79$ & $0.86 \pm 0.09$ & $0.86 \pm 0.79$ & $77.07\pm 0.09$ & $2558711.16 \pm 19816.16$ \\
\hline 
\hline

\multicolumn{1}{|c|}{\multirow{3}{*}{\rotatebox[origin=c]{90}{\textbf{$f_8$\color[HTML]{FFFFFF}}}}} & 
k-BBBC   & $\expnumber{1.70}{+1} \pm \expnumber{6.87}{+0}$ & $\expnumber{2.54}{+1} \pm \expnumber{1.02}{+1}$ & $7.04  \pm 0.53$ & $0.88 \pm 0.07$ & $0.89 \pm 0.05$ & $5.01\pm 0.23$ & $320000 \pm 0.00$ \\
\multicolumn{1}{|c|}{} & 
E-k-BBBC   & $\expnumber{2.00}{+1} \pm \expnumber{1.02}{+1}$ & $\expnumber{2.99}{+1} \pm \expnumber{1.53}{+1}$ & $7.16  \pm 0.78$ & $0.90 \pm 0.10$ & $0.91 \pm 0.07$ & $4.29\pm 0.14$ & $320000 \pm 0.00$ \\
\multicolumn{1}{|c|}{} &  
MCS   & $\expnumber{1.52}{+1} \pm \expnumber{3.86}{-3}$ & $\expnumber{2.61}{+3} \pm \expnumber{5.01}{-6}$ & $4.00 \pm 0.00$ & $0.50 \pm 0.00$ & $1.00 \pm 0.00$ & $5.95\pm 0.58$ & $138391.76 \pm 4403.18$ \\
\hline 
\hline

\multicolumn{1}{|c|}{\multirow{3}{*}{\rotatebox[origin=c]{90}{\textbf{$f_9$\color[HTML]{FFFFFF}}}}} & 
k-BBBC   & $\expnumber{1.71}{+1} \pm \expnumber{1.48}{+0}$ & $\expnumber{8.21}{+0} \pm \expnumber{6.79}{-1}$ & $62.76  \pm 0.90$ & $0.98 \pm 0.01$ & $0.95 \pm 0.03$ & $135.28\pm 2.61$ & $5120000 \pm 0.00$ \\
\multicolumn{1}{|c|}{} & 
E-k-BBBC   & $\expnumber{2.19}{+1} \pm \expnumber{1.01}{+0}$ & $\expnumber{8.34}{+0} \pm \expnumber{3.53}{-1}$ & $62.64  \pm 0.84$ & $0.98 \pm 0.01$ & $0.94 \pm 0.03$ & $132.34\pm 2.38$ & $5120000 \pm 0.00$ \\
\multicolumn{1}{|c|}{} &
MCS   & $\expnumber{6.65}{+1} \pm \expnumber{2.79}{+0}$ & $\expnumber{8.32}{+2} \pm \expnumber{1.50}{+1}$ & $6.96  \pm 0.20$ & $0.11 \pm 0.01$ & $1.00 \pm 0.00$ & $132.12\pm 0.94$ & $1638734.64  \pm 146919.75$ \\
\hline 
\hline

\multicolumn{1}{|c|}{\multirow{3}{*}{\rotatebox[origin=c]{90}{\textbf{$f_{10}$\color[HTML]{FFFFFF}}}}} & 
k-BBBC   & $\expnumber{6.13}{-3} \pm \expnumber{1.70}{-3}$ & $\expnumber{3.98}{-4} \pm \expnumber{2.03}{-4}$ & $4.00  \pm 0.00$ & $1.00 \pm 0.00$ & $1.00 \pm 0.00$ & $4.60\pm 0.16$ & $320000 \pm 0.00$ \\
\multicolumn{1}{|c|}{} & 
E-k-BBBC   & $\expnumber{3.25}{-4} \pm \expnumber{9.23}{-5}$ & $\expnumber{1.06}{-6} \pm \expnumber{5.73}{-7}$ & $4.00  \pm 0.00$ & $1.00 \pm 0.00$ & $1.00 \pm 0.00$ & $4.25\pm 0.05$ & $320000 \pm 0.00$ \\
\multicolumn{1}{|c|}{} &  
MCS   & $\expnumber{3.18}{-3} \pm \expnumber{1.13}{-3}$ & $\expnumber{1.12}{-4} \pm \expnumber{6.12}{-5}$ & $3.72  \pm 0.45$ & $0.93 \pm 0.11$ & $1.00 \pm 0.00$ & $4.62\pm 0.01$ & $165891.64 \pm 7415.37$ \\
\hline 
\hline

\multicolumn{1}{|c|}{\multirow{3}{*}{\rotatebox[origin=c]{90}{\textbf{$f_{11}$\color[HTML]{FFFFFF}}}}} & 
k-BBBC   & $\expnumber{6.26}{-3} \pm \expnumber{1.98}{-3}$ & $\expnumber{1.94}{-3} \pm \expnumber{1.15}{-3}$ & $3.64  \pm 0.48$ & $0.91 \pm 0.12$ & $0.98 \pm 0.07$ & $4.45\pm 0.48$ & $320000 \pm 0.00$ \\
\multicolumn{1}{|c|}{} & 
E-k-BBBC   & $\expnumber{3.87}{-4} \pm \expnumber{1.26}{-4}$ & $\expnumber{1.04}{-4} \pm \expnumber{1.78}{-5}$ & $3.64  \pm 0.56$ & $0.91 \pm 0.14$ & $0.99 \pm 0.05$ & $4.63\pm 0.26$ & $320000 \pm 0.00$ \\ 
\multicolumn{1}{|c|}{} &  
MCS   & $\expnumber{1.65}{-3} \pm \expnumber{6.33}{-4}$ & $\expnumber{1.47}{-4} \pm \expnumber{1.40}{-4}$ & $3.44  \pm 0.58$ & $0.86 \pm 0.14$ & $1.00 \pm 0.00$ & $4.48\pm 0.01$ & $205961.20 \pm 6971.04$ \\ 
\hline 
\hline

\multicolumn{1}{|c|}{\multirow{3}{*}{\rotatebox[origin=c]{90}{\textbf{$f_{12}$\color[HTML]{FFFFFF}}}}} & 
k-BBBC   & $\expnumber{1.49}{-1} \pm \expnumber{8.92}{-3}$ & $\expnumber{4.78}{-2} \pm \expnumber{5.48}{-3}$ & $121.00  \pm 0.00$ & $1.00 \pm 0.00$ & $1.00 \pm 0.00$ & $394.99 \pm 12.29$ &  $9680000 \pm 0.00$ \\
\multicolumn{1}{|c|}{} & 
E-k-BBBC   & $\expnumber{7.71}{-3} \pm \expnumber{3.24}{-4}$ & $\expnumber{1.23}{-4} \pm \expnumber{9.76}{-6}$ & $121.00  \pm 0.00$ & $1.00 \pm 0.00$ & $1.00 \pm 0.00$ & $384.64\pm 11.98$ & $9680000 \pm 0.00$ \\
\multicolumn{1}{|c|}{} &  
MCS   & $\expnumber{5.91}{-2} \pm \expnumber{4.48}{-3}$ & $\expnumber{1.17}{-2} \pm \expnumber{3.17}{-3}$ & $120.96  \pm 0.20$ & $0.99 \pm 0.01$ & $0.99 \pm 0.01$ & $398.94\pm 2.69$ & $2173487.72 \pm 17194.34$ \\
\hline 
\hline

\multicolumn{1}{|c|}{\multirow{3}{*}{\rotatebox[origin=c]{90}{\textbf{$f_{13}$\color[HTML]{FFFFFF}}}}} & 
k-BBBC  & $\expnumber{6.10}{-3} \pm \expnumber{7.98}{-4}$ & $\expnumber{3.37}{-5} \pm \expnumber{9.56}{-6}$ & $24.80  \pm 0.40$ & $0.99 \pm 0.02$ & $0.99 \pm 0.01$ & $30.03 \pm 0.65$ &  $2000000 \pm 0.00$ \\
\multicolumn{1}{|c|}{} & 
E-k-BBBC & $\expnumber{3.05}{-4} \pm \expnumber{3.02}{-5}$ & $\expnumber{7.93}{-8} \pm \expnumber{1.55}{-8}$ & $24.64  \pm 0.48$ & $0.99 \pm 0.02$ & $0.99 \pm 0.01$ & $30.17 \pm 0.72$ &  $2000000 \pm 0.00$ \\
\multicolumn{1}{|c|}{} &  
MCS & $\expnumber{2.80}{-3} \pm \expnumber{4.48}{-4}$ & $\expnumber{1.05}{-5} \pm \expnumber{5.53}{-6}$ & $23.20  \pm 1.17$ & $0.93 \pm 0.05$ & $1.00 \pm 0.00$ & $30.28 \pm 0.16$ &  $567374.24 \pm 7436.09$ \\
\hline 
\hline

\multicolumn{1}{|c|}{\multirow{3}{*}{\rotatebox[origin=c]{90}{\textbf{$f_{14}$\color[HTML]{FFFFFF}}}}} & 
k-BBBC   & $\expnumber{8.17}{-2} \pm \expnumber{1.00}{-2}$ & $\expnumber{2.39}{-5} \pm \expnumber{5.79}{-6}$ & $36.00  \pm 0.00$ & $1.00 \pm 0.00$ & $1.00 \pm 0.00$ & $45.75 \pm 0.78$ &  $2959920 \pm 0.00$ \\
\multicolumn{1}{|c|}{} & 
E-k-BBBC   & $\expnumber{4.32}{-3} \pm \expnumber{3.50}{-4}$ & $\expnumber{6.99}{-8} \pm \expnumber{1.03}{-8}$ & $36.00  \pm 0.00$ & $1.00 \pm 0.00$ & $1.00 \pm 0.00$ & $47.50 \pm 0.67$ &  $2959920 \pm 0.00$ \\
\multicolumn{1}{|c|}{} &  
MCS   & $\expnumber{3.40}{-2} \pm \expnumber{4.43}{-3}$ & $\expnumber{4.35}{-6} \pm \expnumber{1.32}{-6}$ & $35.80  \pm 0.40$ & $0.99 \pm 0.01$ & $1.00 \pm 0.00$ & $46.29 \pm 0.37$ &  $633959.96 \pm 10290.21$ \\
\hline 
\hline

\multicolumn{1}{|c|}{\multirow{3}{*}{\rotatebox[origin=c]{90}{\textbf{$f_{15}$\color[HTML]{FFFFFF}}}}} & 
k-BBBC   & $\expnumber{4.33}{-2} \pm \expnumber{6.99}{-3}$ & $\expnumber{1.47}{-3} \pm \expnumber{5.69}{-4}$ & $25.92  \pm 0.27$ & $0.72 \pm 0.00$ & $0.93 \pm 0.05$ & $46.61 \pm 0.75$ &  $2959920 \pm 0.00$ \\
\multicolumn{1}{|c|}{} & 
E-k-BBBC &  $\expnumber{2.27}{-3} \pm \expnumber{3.10}{-4}$ & $\expnumber{3.35}{-6} \pm \expnumber{9.38}{-7}$ & $26.00  \pm 10.00$ & $0.72 \pm 0.00$ & $0.97 \pm 0.02$ & $47.00 \pm 0.76$ &  $2959920 \pm 0.00$ \\ 
\multicolumn{1}{|c|}{} &  
MCS   & $\expnumber{1.63}{-2} \pm \expnumber{7.62}{-3}$ & $\expnumber{2.02}{-4} \pm \expnumber{1.49}{-4}$ & $19.08  \pm 1.26$ & $0.53 \pm 0.01$ & $1.00 \pm 0.00$ & $50.08 \pm 0.26$ &  $1416016.16 \pm 33656.33$ \\ 
\hline 
\hline

\multicolumn{1}{|c|}{\multirow{3}{*}{\rotatebox[origin=c]{90}{\textbf{$f_{16}$\color[HTML]{FFFFFF}}}}} & 
k-BBBC   & $\expnumber{1.16}{-1} \pm \expnumber{9.49}{-3}$ & $\expnumber{1.27}{-3} \pm \expnumber{2.52}{-4}$ & $55.92  \pm 0.27$ & $0.99 \pm 0.00$ & $0.99 \pm 0.00 $ & $93.37 \pm 4.63$ &  $4480000 \pm 0.00$ \\ 
\multicolumn{1}{|c|}{} & 
E-k-BBBC   & $\expnumber{5.19}{-3} \pm \expnumber{3.52}{-4}$ & $\expnumber{2.71}{-5} \pm \expnumber{7.16}{-7}$ & $55.92  \pm 0.27$ & $0.99 \pm 0.00$ & $0.99 \pm 0.00 $ & $104.98 \pm 1.20$ &  $4480000 \pm 0.00$ \\ 
\multicolumn{1}{|c|}{} &  
MCS   & $\expnumber{6.03}{-3} \pm \expnumber{1.51}{-3}$ & $\expnumber{4.65}{-5} \pm \expnumber{1.48}{-5}$ & $24.00  \pm 0.00$ & $0.43 \pm 0.00$ & $0.39 \pm 0.00 $ & $93.94 \pm 0.35$ &  $1629539 \pm 19329.97$ \\ 
\hline 
\hline

\multicolumn{1}{|c|}{\multirow{3}{*}{\rotatebox[origin=c]{90}{\textbf{$f_{17}$\color[HTML]{FFFFFF}}}}} & 
k-BBBC   & $\expnumber{1.48}{+0} \pm \expnumber{0.65}{+0}$ & $\expnumber{2.51}{-4} \pm \expnumber{1.50}{-4}$ & $49.00  \pm 0.00$ & $1.00 \pm 0.00$ & $1.00 \pm 0.00$ & $79.92 \pm 1.48$ &  $3920000 \pm 0.00$ \\
\multicolumn{1}{|c|}{} & 
E-k-BBBC   & $\expnumber{1.28}{+0} \pm \expnumber{0.58}{+0}$ & $\expnumber{1.29}{-5} \pm \expnumber{8.52}{-4}$ & $49.00  \pm 0.00$ & $1.00 \pm 0.00$ & $1.00 \pm 0.00$ & $83.84 \pm 1.68$ &  $3920000 \pm 0.00$ \\
\multicolumn{1}{|c|}{} &  
MCS   & $\expnumber{1.29}{+0} \pm \expnumber{0.60}{+0}$ & $\expnumber{2.05}{-4} \pm \expnumber{1.10}{-4}$ & $49.00  \pm 0.00$ & $1.00 \pm 0.00$ & $1.00 \pm 0.00$ & $80.53 \pm 0.47$ &  $858546.76 \pm 11423.67$ \\
\hline 
\hline

\multicolumn{1}{|c|}{\multirow{3}{*}{\rotatebox[origin=c]{90}{\textbf{$f_{18}$\color[HTML]{FFFFFF}}}}} & 
k-BBBC   & $\expnumber{1.12}{-2} \pm \expnumber{1.50}{-3}$ & $\expnumber{4.47}{-4} \pm \expnumber{1.28}{-4}$ & $9.00  \pm 0.00$ & $1.00 \pm 0.00$ & $1.00 \pm 0.00$ & $8.77 \pm \expnumber{9.95}{-2}$ &  $720000 \pm 0.00$ \\ 
\multicolumn{1}{|c|}{} & 
E-k-BBBC   & $\expnumber{5.70}{-4} \pm \expnumber{1.01}{-4}$ & $\expnumber{1.15}{-6} \pm \expnumber{3.95}{-7}$ & $9.00  \pm 0.00$ & $1.00 \pm 0.00$ & $1.00 \pm 0.00$ & $8.76 \pm \expnumber{1.71}{-1}$ & $720000 \pm 0.00$ \\ 
\multicolumn{1}{|c|}{} &  
MCS   & $\expnumber{4.91}{-3} \pm \expnumber{9.91}{-4}$ & $\expnumber{1.22}{-4} \pm \expnumber{8.06}{-5}$ & $8.96  \pm 0.20$ & $0.99 \pm 0.02$ & $1.00 \pm 0.00$ & $8.82 \pm \expnumber{3.09}{-2}$ &  $337034.72 \pm 11448.13$ \\ 
\hline 
\hline

\multicolumn{1}{|c|}{\multirow{3}{*}{\rotatebox[origin=c]{90}{\textbf{$f_{19}$\color[HTML]{FFFFFF}}}}} & 
k-BBBC   & $\expnumber{4.50}{+0} \pm \expnumber{0.16}{+0}$ & $\expnumber{0.02}{+0} \pm \expnumber{1.64}{-3}$ & $379.00  \pm 0.00$ & $1.00 \pm 0.00$ & $1.00 \pm 0.00$ & $2832.24 \pm 225.21$ &  $30320000 \pm 0.00$ \\ 
\multicolumn{1}{|c|}{} & 
E-k-BBBC   & $\expnumber{0.23}{+0} \pm \expnumber{6.75}{-3}$ & $\expnumber{7.14}{-5} \pm \expnumber{3.28}{-6}$ & $379.00  \pm 0.00$ & $1.00 \pm 0.00$ & $1.00 \pm 0.00$ & $3159.67 \pm 124.68$ &  $30320000 \pm 0.00$ \\  
\multicolumn{1}{|c|}{} &  
MCS    & $\expnumber{1.01}{+0} \pm \expnumber{4.08}{-2}$ & $\expnumber{1.64}{-3} \pm \expnumber{1.62}{-4}$ & $379.00  \pm 0.00$ & $1.00 \pm 0.00$ & $1.00 \pm 0.00$ & $2857.54 \pm 15.05$ &  $6545999.12 \pm 85841.39$ \\  
\hline 
\hline

\multicolumn{1}{|c|}{\multirow{3}{*}{\rotatebox[origin=c]{90}{\textbf{$f_{20}$\color[HTML]{FFFFFF}}}}} & 
k-BBBC   & $\expnumber{4.51}{+0} \pm \expnumber{0.10}{+0}$ & $\expnumber{0.02}{+0} \pm \expnumber{1.08}{-3}$ & $379.00  \pm 0.00$ & $1.00 \pm 0.00$ & $1.00 \pm 0.00$ & $3031.97 \pm 1115.85$ &  $30320000 \pm 0.00$ \\ 
\multicolumn{1}{|c|}{} & 
E-k-BBBC   & $\expnumber{0.23}{+0} \pm \expnumber{5.01}{-3}$ & $\expnumber{6.96}{-5} \pm \expnumber{2.39}{-6}$ & $379.00  \pm 0.00$ & $1.00 \pm 0.00$ & $1.00 \pm 0.00$ & $3550.65 \pm 191.99$ &  $30320000 \pm 0.00$ \\ 
\multicolumn{1}{|c|}{} &  
MCS   & $\expnumber{1.02}{+0} \pm \expnumber{0.05}{+0}$ & $\expnumber{1.70}{-3} \pm \expnumber{1.82}{-4}$ & $379.00  \pm 0.00$ & $1.00 \pm 0.00$ & $1.00 \pm 0.00$ & $3058.77 \pm 19.89$ &  $6685027.92 \pm 307249.28$ \\ 
\hline 

\end{tabular}%
}
\end{table*}
\begin{table*}[h!]
\captionsetup{justification=centering}
\caption{Results for High Dimensionality Problems}
\label{tab:results_high}
\resizebox{\linewidth}{!}{%
\begin{tabular}{c|c|c|c|c|c|c|c|c|}
\cline{2-9}
\multicolumn{1}{l|}{}                                   & \makecell{\textbf{Algorithm}} & \makecell{\textbf{Accuracy} \\ \textbf{Search Space} \\ $a_{src}$}           & \makecell{\textbf{Accuracy} \\ \textbf{Objective Space} \\ $a_{obj}$}   & \makecell{\textbf{Detected} \\ \textbf{Local Optima} \\ $\bar{m}$}    & \makecell{ \textbf{Success Rate} \\ Actual \\ $\sfrac{\bar{m}}{m}$ } & \makecell{ \textbf{Success Rate} \\ Quantification \\ $\sfrac{m_c}{m}$}  & \makecell{\textbf{Runtime} \\ $\left[\text{seconds}\right]$}   & \makecell{\textbf{Num. of Function} \\ \textbf{Evaluations}}
 \\ \hline

\multicolumn{1}{|c|}{\multirow{3}{*}{\rotatebox[origin=c]{90}{\textbf{$d=4$\color[HTML]{FFFFFF}}}}} & 
k-BBBC   & $\expnumber{1.88}{-2} \pm \expnumber{1.33}{-3}$ & $\expnumber{9.83}{-3} \pm \expnumber{9.88}{-4}$ & $48.00  \pm 0.00$ & $1.00 \pm 0.00$ & $1.00 \pm 0.00$ & $265.24 \pm 4.79$ &  $7680000 \pm 0.00$ \\ 
\multicolumn{1}{|c|}{} & 
E-k-BBBC   & $\expnumber{4.75}{-3} \pm \expnumber{1.79}{-4}$ & $\expnumber{8.65}{-4} \pm \expnumber{3.89}{-5}$ & $48.00  \pm 0.00$ & $1.00 \pm 0.00$ & $1.00 \pm 0.00$ & $348.80\pm 3.11$ &  $7680000 \pm 0.00$ \\ 
\multicolumn{1}{|c|}{} &  
MCS   & $\expnumber{1.72}{-1} \pm \expnumber{1.30}{-2}$ & $\expnumber{9.53}{-1} \pm \expnumber{1.42}{-1}$ & $44.84  \pm 1.59$ & $0.93 \pm 0.03$ & $1.00 \pm 0.00$ & $269.59\pm 2.49$ &  $2242255.72 \pm 62569.82$ \\
\hline 
\hline

\multicolumn{1}{|c|}{\multirow{3}{*}{\rotatebox[origin=c]{90}{\textbf{$d=8$\color[HTML]{FFFFFF}}}}} & 
k-BBBC   & $\expnumber{5.65}{-2} \pm \expnumber{1.81}{-3}$ & $\expnumber{2.60}{-2} \pm \expnumber{1.27}{-3}$ & $48.00  \pm 0.00$ & $1.00 \pm 0.00$ & $1.00 \pm 0.00$ & $1067.34 \pm  13.37$ &  $15360000 \pm 0.00$ \\ 
\multicolumn{1}{|c|}{} & 
E-k-BBBC   & $\expnumber{2.88}{-2} \pm \expnumber{8.33}{-4}$ & $\expnumber{6.59}{-3} \pm \expnumber{2.47}{-4}$ & $48.00  \pm 0.00$ & $1.00 \pm 0.00$ & $1.00 \pm 0.00$ & $1194.95 \pm  65.59$ &  $15360000 \pm 0.00$  \\ 
\multicolumn{1}{|c|}{} &  
MCS   & $-$ & $-$ & $0.00 \pm 0.00$ & $0.00 \pm 0.00$ & $1.00 \pm 0.00$ & $1092.92 \pm 14.94$ & $1698917.48 \pm 44782.16$ \\
\hline 
\hline

\multicolumn{1}{|c|}{\multirow{3}{*}{\rotatebox[origin=c]{90}{\textbf{$d=16$\color[HTML]{FFFFFF}}}}} & 
k-BBBC   & $\expnumber{1.15}{-1} \pm \expnumber{2.22}{-3}$ & $\expnumber{7.45}{-2} \pm \expnumber{1.87}{-3}$ & $48.00  \pm 0.00$ & $1.00 \pm 0.00$ & $1.00 \pm 0.00$ & $4987.12 \pm  293.58$ &  $30720000 \pm 0.00$ \\ 
\multicolumn{1}{|c|}{} & 
E-k-BBBC   & $\expnumber{7.50}{-2} \pm \expnumber{1.19}{-3}$ & $\expnumber{3.09}{-2} \pm \expnumber{7.67}{-4}$ & $48.00  \pm 0.00$ & $1.00 \pm 0.00$ & $1.00 \pm 0.00$ & $5313.94 \pm  101.93$ &  $30720000 \pm 0.00$ \\ 
\multicolumn{1}{|c|}{} &  
MCS   & $-$ & $-$ & $0.00 \pm 0.00$ & $0.00 \pm 0.00$ & $1.00 \pm 0.00$ & $5096.13 \pm 67.82$ & $6986781.24 \pm 610442.82$ \\
\hline 
\hline

\multicolumn{1}{|c|}{\multirow{3}{*}{\rotatebox[origin=c]{90}{\textbf{$d=32$\color[HTML]{FFFFFF}}}}} & 
k-BBBC   & $\expnumber{2.58}{-1} \pm \expnumber{2.61}{-3}$ & $\expnumber{2.33}{-1} \pm \expnumber{5.60}{-3}$ & $47.04  \pm 0.49$ & $0.99 \pm 0.01$ & $0.99 \pm 0.01$ & $38678.88 \pm  722.39$ &  $61440000 \pm 0.00$ \\ 
\multicolumn{1}{|c|}{} & 
E-k-BBBC   & $\expnumber{2.08}{-1} \pm \expnumber{6.90}{-4}$ & $\expnumber{2.41}{-1} \pm \expnumber{2.65}{-3}$ & $48.00  \pm 0.00$ & $1.00 \pm 0.00$ & $1.00 \pm 0.00$ & $40063.19 \pm  689.15$ &  $61440000 \pm 0.00$  \\ 
\multicolumn{1}{|c|}{} &  
MCS   & $-$ & $-$ & $-$ & $-$ & $-$ & $-$ & $-$ \\
\hline 

\end{tabular}%
}
\end{table*}

\subsection{Experiments with high dimensionality problems}
\label{sec:high_dim_experiments}

We tested k-BBBC on the high-dimensional multimodal test problem defined in Eqn.~(\ref{eq:debMMP}). This function is similar to Rastrigin’s function and it is a modified version of the Key function defined in Eqn.~(\ref{eq:key}). In the range $\left[0,1\right]$ and with the $j$ values defined as in Eqn.~(\ref{eq:debMMP}), it features $48$ local minima (one of them being global) -- the exact optima values $\textbf{z}$ are reported in the work of Deb and Saha~\cite{deb2012multimodal}. We tested the algorithms for different dimensionalities: $d = 4$, $8$, $16$, and $32$. Due to the multi-dimensionality of the function, we will not report its graph.

\begin{equation}
    \label{eq:debMMP}
    \begin{aligned}
    \textrm{minimize}  \quad & f(\textbf{x}) = \sum_{i=1}^{d}{10 (1+\cos{2 \pi j_i x_i}) + 2 j_i x_i^2} \\
    \textrm{subject to}         \quad & 0 \leq x_i \leq 1, \quad i=1,2,...,d \\
    \end{aligned}
\end{equation}
$$ j_{\sfrac{d}{4}}=2 \quad j_{\sfrac{d}{2}}=2 \quad j_{d\sfrac{3}{4}}=3 \quad j_d=4 $$
$$ j_i = 1 \quad \forall i \neq \sfrac{d}{4}, \sfrac{d}{2}, d\sfrac{3}{4}, d \\ $$

While we did not directly compare k-BBBC with Deb and Saha's method~\cite{deb2012multimodal}, the authors claim that their method could retrieve all optima for the problem up to $d=16$. This will be our experimental baseline.

\begin{figure*}[t!]
    \centering
    \subfigure[\protect\url{}\label{fig:accuracySS_total} Accuracy (Search Space)]
    {\includegraphics[height=5cm]{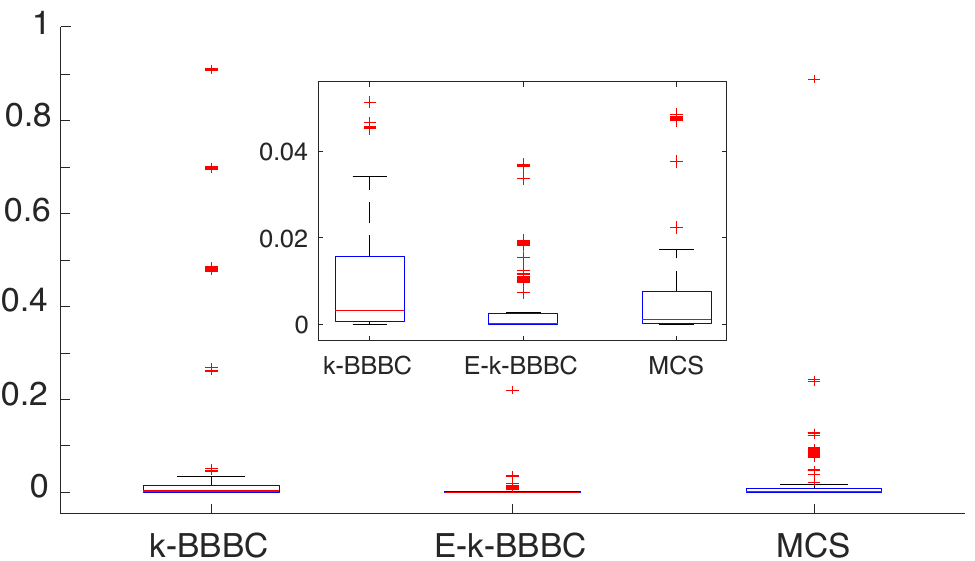}} 
    \subfigure[\protect\url{}\label{fig:success_rate_total} Success Rate] 
    {\includegraphics[height=5cm]{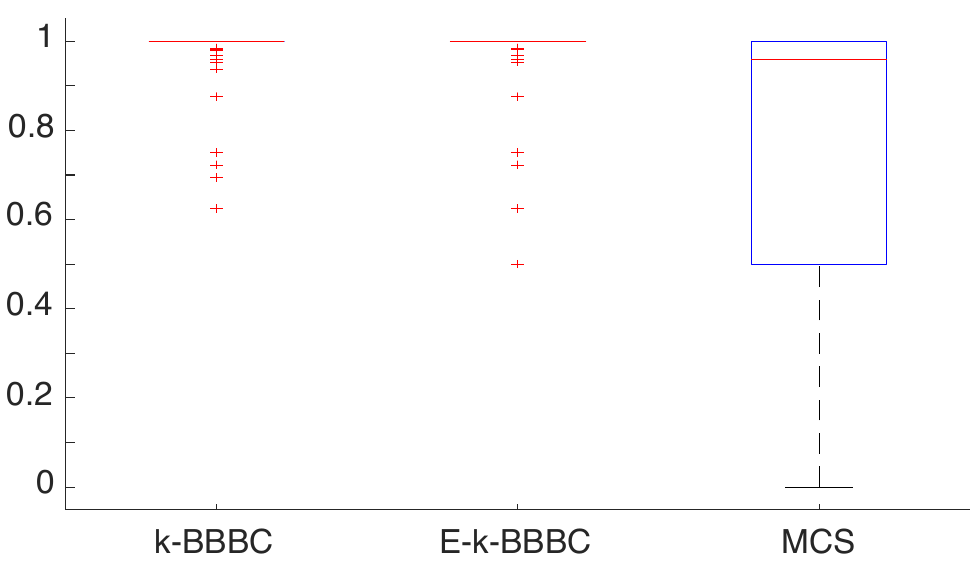}}
    \caption{Result comparisons among the algorithms over every problem. Plots include median, interquartile range, and outliers. The plot in (a) reports the accuracy in search space $a_{src}$ normalized over each function's domain; note that these results do not include incomplete runs or with a success rate equal to $0$. The inner plot zooms in a smaller interval of accuracy to appreciate differences without including some outliers. The plot in (b) reports the (actual) success rate $\sfrac{\bar{m}}{m}$ of each algorithm; note that incomplete runs have been given a success rate equal to $0$.}
    \label{fig:box_whisk}
\end{figure*}

\subsection{Results}
\label{sec:results}

Table~\ref{tab:results_low} and~\ref{tab:results_high} show the results for problems having low and high dimensionality, respectively -- reporting average and standard deviation for each metric. 

The plot in Figure~\ref{fig:accuracySS_total} reports the overall accuracy in search space for each algorithm, and each result was normalized against their function's domain. The plot evidences a higher accuracy of elitist k-BBBC, outperforming both k-BBBC and MCS. Note that the accuracy in objective space was not reported as a plot due to the challenge of normalizing values against the codomain of each problem. Additionally, Table~\ref{tab:results_high} shows that k-BBBC and its elitist version feature a higher accuracy (in both search and objective space) compared to MCS; more importantly, MCS could not find any optima for problems with $d\geq8$, and could not finish its execution with $d=32$ due to extensive computational requests, suggesting that k-BBBC can tackle high dimensionality problems better than other algorithms.

Figure~\ref{fig:success_rate_total} summarizes the actual success rate of each algorithm (i.e., retrieved solutions compared against the actual local optima). Although MCS performs fairly well in most of the problems, it is outperformed by both versions of k-BBBC -- this is evident in $f_6$, $f_{8}$, $f_{9}$, $f_{15}$, and $f_{16}$. Furthermore, as previously mentioned, MCS could not find any optima for functions with $d\geq8$, resulting in a success rate equal to $0$.

The success rate retrieved from the quantification is consistent with the actual one, but only for k-BBBC (elitist and non). For MCS, although the quantification was mostly consistent, we can spot some imperfections (e.g., $f_6$, $f_9$, $f_{10}$, $f_{11}$, and $f_{15}$); this is mostly due to the population not converging entirely to the local optima (i.e., not forming clusters), which prevents the quantification to be applied correctly -- the most exacerbated case being $f_9$ where the quantification reported a correct retrieval of all optima, whereas in truth MCS only retrieved $7$ over $64$ optima; this is because the accuracy in search space was not sufficiently high to match every retrieved optima with the actual ones. The only exception where the quantification method failed also for k-BBBC is $f_{15}$ (Vincent's function), where the size of the concave downwards/upwards in the search space between local optima is progressively increasing with the domain, deceiving clustering algorithms due to the high concentration of optima in a specific region. This concentration leads the clustering algorithm to group all the points within the same cluster, and since the distances within this region are smaller than the \textit{incorrect} points splitting a single optimum, the algorithm fails. 

As the runtime was set to be consistent for all algorithms, we also reported the number of function evaluations during each run. Both k-BBBC and its elitist version feature the same number of evaluations, which is constant throughout all the runs as individuals are only evaluated once per generation (note that the elites are not re-evaluated in following generations); on the other hand, MCS features a variable number of evaluations due to its depuration procedure. All algorithms show a similar number of evaluations, with MCS generally performing fewer evaluations depending on when the depuration procedure is triggered; this, however, might affect the results of the optimization -- i.e., if the run is terminated abruptly, MCS might not have had the time to retrieve neighbors and more optimal solutions. This might also suggest an alternative observation, which is MCS has a higher time complexity than k-BBBC (although the authors did not report any analysis in the original work~\cite{cuevas2014cuckoo}), leading to a termination that could also happen midrun and therefore preventing the algorithm to converge to all local optima.

\section{Experiments for Finding Global Peaks (on CEC 2013 Benchmark set)}
\label{sec:experiments_global}

Multimodal optimization algorithms are usually tested on the widely accepted CEC'2013 test suite for niching competition~\cite{li2013benchmark}. However, this benchmark set is specifically designed for retrieving \textit{only} global optima (namely, the peaks in the landscape of the optimization problem), ignoring any local peak. k-BBBC is not designed for this purpose, as it aims to find every optimum (either global or local) within the problem, which requires considerable computational resources when this number is large -- like in the last problems of CEC'2013. From the experiment results in Sec.~\ref{sec:experiments_local}, we expect k-BBBC to perform well when the problem has few local optima and badly when there are many. Furthermore, k-BBBC does not feature any niching strategy, which creates a disadvantage in functions with irregularly distributed optima (i.e., Vincent function). Nevertheless, we tested E-k-BBBC on this benchmark for two reasons: (i) to understand how it performs on problems for which it was not designed and evaluate how far BBBC is from a promising path in large-scale multimodal optimization, and (ii) to compare it with the latest state-of-the-art methods since the results are publicly available, and there was no need to implement them as in Sec.~\ref{sec:experiments_local}.

\subsection{Benchmark Functions and Performance Metrics} \label{sec:expG_benchmark_metrics}

We evaluated the performance of k-BBBC (specifically, its elitist version) using the CEC'2013 benchmark set~\cite{li2013benchmark}, which consists of twenty multimodal maximization problems. These problems, along with their characteristics, are detailed in Table~\ref{tab:functions_CEC13}. Specifically, $F_1-F_3$ are simple one-dimensional functions, $F_4-F_7$ and $F_{10}$ are scalable two-dimensional functions, $F_8-F_9$ are scalable three-dimensional functions, and $F_{11}-F_{20}$ are complex composition functions with dimensions ranging from two to twenty.

\begin{table*}[]
\centering
\caption{Properties of the twenty multimodal test functions in the CEC'2013 benchmark set}
\label{tab:functions_CEC13}
\resizebox{\linewidth}{!}{%
\begin{tabular}{|c|c|c|c|c|c|}
\hline
\textbf{Function} & \textbf{Name} & \textbf{Dimension Size} & \textbf{No. of Global Optima} & \textbf{No. of Local Optima} & \textbf{Max Function Evaluations} \\ \hline
$F_{1}$ & Five-Uneven-Peak Trap & 1 & 2 & 3 & $\expnumber{5.0}{+04}$ \\ \hline
$F_{2}$ & Equal Maxima & 1 & 5 & 0 & $\expnumber{5.0}{+04}$ \\ \hline
$F_{3}$ & Uneven Decreasing Maxima & 1 & 1 & 4 & $\expnumber{5.0}{+04}$ \\ \hline
$F_{4}$ & Himmelblau & 2 & 4 & 0 & $\expnumber{5.0}{+04}$ \\ \hline
$F_{5}$ & Six-Hump Camel Back & 2 & 2 & 2 & $\expnumber{5.0}{+04}$ \\ \hline
$F_{6}$ & Shubert with 2D & 2 & 18 & many & $\expnumber{2.0}{+05}$ \\ \hline
$F_{7}$ & Vincent with 2D & 2 & 36 & 0 & $\expnumber{2.0}{+05}$ \\ \hline
$F_{8}$ & Shubert with 3D & 3 & 81 & many & $\expnumber{4.0}{+05}$ \\ \hline
$F_{9}$ & Vincent with 3D & 3 & 216 & 0 & $\expnumber{4.0}{+05}$ \\ \hline
$F_{10}$ & Modified Rastrigin & 2 & 12 & 0 & $\expnumber{2.0}{+05}$ \\ \hline
$F_{11}$ & Composition Function 1 with 2D & 2 & 6 & many & $\expnumber{2.0}{+05}$ \\ \hline
$F_{12}$ & Composition Function 2 with 2D & 2 & 8 & many & $\expnumber{2.0}{+05}$ \\ \hline
$F_{13}$ & Composition Function 3 with 2D & 2 & 6 & many & $\expnumber{2.0}{+05}$ \\ \hline
$F_{14}$ & Composition Function 3 with 3D & 3 & 6 & many & $\expnumber{4.0}{+05}$ \\ \hline
$F_{15}$ & Composition Function 4 with 3D & 3 & 8 & many & $\expnumber{4.0}{+05}$ \\ \hline
$F_{16}$ & Composition Function 3 with 5D & 5 & 6 & many & $\expnumber{4.0}{+05}$ \\ \hline
$F_{17}$ & Composition Function 4 with 5D & 5 & 8 & many & $\expnumber{4.0}{+05}$ \\ \hline
$F_{18}$ & Composition Function 3 with 10D & 10 & 6 & many & $\expnumber{4.0}{+05}$ \\ \hline
$F_{19}$ & Composition Function 4 with 10D & 10 & 8 & many & $\expnumber{4.0}{+05}$ \\ \hline
$F_{20}$ & Composition Function 4 with 20D & 20 & 8 & many & $\expnumber{4.0}{+05}$ \\ \hline
\end{tabular}%
}
\end{table*}

We solved each problem by setting a maximum number of function evaluations (MaxFEs), following the procedure outlined in the literature~\cite{li2013benchmark}, as indicated in the rightmost column of Table~\ref{tab:functions_CEC13}. MaxFEs determines the total number of runs (NR) for each problem based on the population size. Using this setup, we assessed the algorithm’s performance with two specific metrics:
\begin{itemize}
    \item \textit{peak ratio (PR)} refers to the average percentage of global peaks identified across multiple runs. As defined in Eqn.~(\ref{eq:pr}), it is calculated by summing the number of global peaks found in each run (NPF), then dividing this by the product of the total number of global peaks (TPN) and the total number of runs (NR); and
    \begin{equation}
        \label{eq:pr}
        \text{PR} = \frac{\sum^{\text{NR}}_{i=1}{\text{NPF}_i}}{\text{TNP} \cdot \text{NR}}
    \end{equation}
    
    \item \textit{success ratio (SR)} represents the percentage of runs that successfully identified all global optima across multiple attempts. As shown in Eqn.~(\ref{eq:sr}), it is calculated as the ratio of successful runs (NSR) to the total number of runs (NR).

    \begin{equation}
        \label{eq:sr}
        \text{SR} = \frac{\text{NSR}}{\text{NR}}
    \end{equation}
    
\end{itemize}
Based on the procedure established in the literature~\cite{wang2017dual, deb2012multimodal}, we analyzed and discussed the results with accuracy $\varepsilon = \expnumber{1.0}{-04}$.

\begin{table*}[]
\centering
\caption{Experimental results of peak ratio and success rate on the CEC'2013 benchmark set at the accuracy level $\varepsilon=1.0E-04$}
\label{tab:comparison_4}
\resizebox{\textwidth}{!}{
\begin{tabular}{|ccc|cc|cc|cc|cc|cc|cc|}
\hline
\multicolumn{1}{|c|}{\multirow{2}{*}{Func.}} & \multicolumn{2}{c|}{E-k-BBBC} & \multicolumn{2}{c|}{SPEA-DE/rand/1} & \multicolumn{2}{c|}{SSGA} & \multicolumn{2}{c|}{CDE} & \multicolumn{2}{c|}{RSCMSA} & \multicolumn{2}{c|}{LBP-ADE} & \multicolumn{2}{c|}{HillVallEA18} \\ \cline{2-15} 
\multicolumn{1}{|c|}{} & \multicolumn{1}{c|}{PR} & SR & \multicolumn{1}{c|}{PR} & SR & \multicolumn{1}{c|}{PR} & SR & \multicolumn{1}{c|}{PR} & SR & \multicolumn{1}{c|}{PR} & SR & \multicolumn{1}{c|}{PR} & SR & \multicolumn{1}{c|}{PR} & SR \\ \hline
\multicolumn{1}{|c|}{$F_{1}$} & \multicolumn{1}{c|}{\textbf{1.000}} & \textbf{1.000} & \multicolumn{1}{c|}{\textbf{1.000($\approx$)}} & \textbf{1.000} & \multicolumn{1}{c|}{\textbf{1.000($\approx$)}} & \textbf{1.000} & \multicolumn{1}{c|}{\textbf{1.000($\approx$)}} & \textbf{1.000} & \multicolumn{1}{c|}{\textbf{1.000($\approx$)}} & \textbf{1.000} & \multicolumn{1}{c|}{\textbf{1.000($\approx$)}} & \textbf{1.000} & \multicolumn{1}{c|}{\textbf{1.000($\approx$)}} & \textbf{1.000} \\ \hline
\multicolumn{1}{|c|}{$F_{2}$} & \multicolumn{1}{c|}{\textbf{1.000}} & \textbf{1.000} & \multicolumn{1}{c|}{\textbf{1.000($\approx$)}} & \textbf{1.000} & \multicolumn{1}{c|}{\textbf{1.000($\approx$)}} & \textbf{1.000} & \multicolumn{1}{c|}{\textbf{1.000($\approx$)}} & \textbf{1.000} & \multicolumn{1}{c|}{\textbf{1.000($\approx$)}} & \textbf{1.000} & \multicolumn{1}{c|}{\textbf{1.000($\approx$)}} & \textbf{1.000} & \multicolumn{1}{c|}{\textbf{1.000($\approx$)}} & \textbf{1.000} \\ \hline
\multicolumn{1}{|c|}{$F_{3}$} & \multicolumn{1}{c|}{\textbf{1.000}} & \textbf{1.000} & \multicolumn{1}{c|}{\textbf{1.000($\approx$)}} & \textbf{1.000} & \multicolumn{1}{c|}{\textbf{1.000($\approx$)}} & \textbf{1.000} & \multicolumn{1}{c|}{\textbf{1.000($\approx$)}} & \textbf{1.000} & \multicolumn{1}{c|}{\textbf{1.000($\approx$)}} & \textbf{1.000} & \multicolumn{1}{c|}{\textbf{1.000($\approx$)}} & \textbf{1.000} & \multicolumn{1}{c|}{\textbf{1.000($\approx$)}} & \textbf{1.000} \\ \hline
\multicolumn{1}{|c|}{$F_{4}$} & \multicolumn{1}{c|}{\textbf{1.000}} & \textbf{1.000} & \multicolumn{1}{c|}{\textbf{1.000($\approx$)}} & \textbf{1.000} & \multicolumn{1}{c|}{0.155($+$)} & 0.000 & \multicolumn{1}{c|}{0.985($+$)} & 0.940 & \multicolumn{1}{c|}{0.995($+$)} & 0.980 & \multicolumn{1}{c|}{\textbf{1.000($\approx$)}} & \textbf{1.000} & \multicolumn{1}{c|}{\textbf{1.000($\approx$)}} & \textbf{1.000} \\ \hline
\multicolumn{1}{|c|}{$F_{5}$} & \multicolumn{1}{c|}{\textbf{1.000}} & \textbf{1.000} & \multicolumn{1}{c|}{\textbf{1.000($\approx$)}} & \textbf{1.000} & \multicolumn{1}{c|}{0.460($+$)} & 0.200 & \multicolumn{1}{c|}{\textbf{1.000($\approx$)}} & \textbf{1.000} & \multicolumn{1}{c|}{0.980($+$)} & 0.960 & \multicolumn{1}{c|}{\textbf{1.000($\approx$)}} & \textbf{1.000} & \multicolumn{1}{c|}{\textbf{1.000($\approx$)}} & \textbf{1.000} \\ \hline
\multicolumn{1}{|c|}{$F_{6}$} & \multicolumn{1}{c|}{0.943} & 0.280 & \multicolumn{1}{c|}{\textbf{1.000($-$)}} & \textbf{1.000} & \multicolumn{1}{c|}{\textbf{1.000($-$)}} & \textbf{1.000} & \multicolumn{1}{c|}{0.640($+$)} & 0.020 & \multicolumn{1}{c|}{0.999($-$)} & 0.980 & \multicolumn{1}{c|}{\textbf{1.000($-$)}} & \textbf{1.000} & \multicolumn{1}{c|}{\textbf{1.000($-$)}} & \textbf{1.000} \\ \hline
\multicolumn{1}{|c|}{$F_{7}$} & \multicolumn{1}{c|}{0.744} & 0.000 & \multicolumn{1}{c|}{0.972($-$)} & 0.240 & \multicolumn{1}{c|}{0.901($-$)} & 0.000 & \multicolumn{1}{c|}{0.719($+$)} & 0.000 & \multicolumn{1}{c|}{0.962($-$)} & 0.280 & \multicolumn{1}{c|}{0.889($-$)} & 0.000 & \multicolumn{1}{c|}{\textbf{1.000($-$)}} & \textbf{1.000} \\ \hline
\multicolumn{1}{|c|}{$F_{8}$} & \multicolumn{1}{c|}{0.000} & 0.000 & \multicolumn{1}{c|}{0.533($-$)} & 0.000 & \multicolumn{1}{c|}{\textbf{1.000($-$)}} & \textbf{1.000} & \multicolumn{1}{c|}{0.530($-$)} & 0.000 & \multicolumn{1}{c|}{0.871($-$)} & 0.000 & \multicolumn{1}{c|}{0.575($-$)} & 0.000 & \multicolumn{1}{c|}{0.920($-$)} & 0.000 \\ \hline
\multicolumn{1}{|c|}{$F_{9}$} & \multicolumn{1}{c|}{0.005} & 0.000 & \multicolumn{1}{c|}{0.723($-$)} & 0.000 & \multicolumn{1}{c|}{0.518($-$)} & 0.000 & \multicolumn{1}{c|}{0.269($-$)} & 0.000 & \multicolumn{1}{c|}{0.627($-$)} & 0.000 & \multicolumn{1}{c|}{0.476($-$)} & 0.000 & \multicolumn{1}{c|}{0.945($-$)} & 0.000 \\ \hline
\multicolumn{1}{|c|}{$F_{10}$} & \multicolumn{1}{c|}{\textbf{1.000}} & \textbf{1.000} & \multicolumn{1}{c|}{\textbf{1.000($\approx$)}} & \textbf{1.000} & \multicolumn{1}{c|}{\textbf{1.000($\approx$)}} & \textbf{1.000} & \multicolumn{1}{c|}{\textbf{1.000($\approx$)}} & \textbf{1.000} & \multicolumn{1}{c|}{\textbf{1.000($\approx$)}} & \textbf{1.000} & \multicolumn{1}{c|}{\textbf{1.000($\approx$)}} & \textbf{1.000} & \multicolumn{1}{c|}{\textbf{1.000($\approx$)}} & \textbf{1.000} \\ \hline
\multicolumn{1}{|c|}{$F_{11}$} & \multicolumn{1}{c|}{\textbf{1.000}} & \textbf{1.000} & \multicolumn{1}{c|}{\textbf{1.000($\approx$)}} & \textbf{1.000} & \multicolumn{1}{c|}{\textbf{1.000($\approx$)}} & \textbf{1.000} & \multicolumn{1}{c|}{0.667($+$)} & 0.000 & \multicolumn{1}{c|}{0.997($+$)} & 0.980 & \multicolumn{1}{c|}{0.674($+$)} & 0.000 & \multicolumn{1}{c|}{\textbf{1.000($\approx$)}} & \textbf{1.000} \\ \hline
\multicolumn{1}{|c|}{$F_{12}$} & \multicolumn{1}{c|}{0.920} & 0.360 & \multicolumn{1}{c|}{\textbf{1.000($-$)}} & \textbf{1.000} & \multicolumn{1}{c|}{\textbf{1.000($-$)}} & \textbf{1.000} & \multicolumn{1}{c|}{0.000($+$)} & 0.000 & \multicolumn{1}{c|}{0.948($-$)} & 0.580 & \multicolumn{1}{c|}{0.750($+$)} & 0.000 & \multicolumn{1}{c|}{\textbf{1.000($-$)}} & \textbf{1.000} \\ \hline
\multicolumn{1}{|c|}{$F_{13}$} & \multicolumn{1}{c|}{0.993} & 0.960 & \multicolumn{1}{c|}{\textbf{1.000($-$)}} & \textbf{1.000} & \multicolumn{1}{c|}{0.957($+$)} & 0.760 & \multicolumn{1}{c|}{0.667($+$)} & 0.000 & \multicolumn{1}{c|}{0.997($-$)} & 0.980 & \multicolumn{1}{c|}{0.667($+$)} & 0.000 & \multicolumn{1}{c|}{\textbf{1.000($-$)}} & \textbf{1.000} \\ \hline
\multicolumn{1}{|c|}{$F_{14}$} & \multicolumn{1}{c|}{0.857} & 0.140 & \multicolumn{1}{c|}{0.839($+$)} & 0.000 & \multicolumn{1}{c|}{0.727($+$)} & 0.020 & \multicolumn{1}{c|}{0.667($+$)} & 0.000 & \multicolumn{1}{c|}{0.810($+$)} & 0.060 & \multicolumn{1}{c|}{0.667($+$)} & 0.000 & \multicolumn{1}{c|}{\textbf{0.917($-$)}} & \textbf{0.560} \\ \hline
\multicolumn{1}{|c|}{$F_{15}$} & \multicolumn{1}{c|}{0.658} & 0.000 & \multicolumn{1}{c|}{\textbf{0.750($-$)}} & 0.000 & \multicolumn{1}{c|}{0.563($+$)} & 0.000 & \multicolumn{1}{c|}{0.528($+$)} & 0.000 & \multicolumn{1}{c|}{0.748($-$)} & 0.000 & \multicolumn{1}{c|}{0.654($+$)} & 0.000 & \multicolumn{1}{c|}{\textbf{0.750($-$)}} & 0.000 \\ \hline
\multicolumn{1}{|c|}{$F_{16}$} & \multicolumn{1}{c|}{\textbf{0.710}} & 0.000 & \multicolumn{1}{c|}{0.663($+$)} & 0.000 & \multicolumn{1}{c|}{0.673($+$)} & 0.000 & \multicolumn{1}{c|}{0.667($+$)} & 0.000 & \multicolumn{1}{c|}{0.667($+$)} & 0.000 & \multicolumn{1}{c|}{0.667($+$)} & 0.000 & \multicolumn{1}{c|}{0.687($+$)} & 0.000 \\ \hline
\multicolumn{1}{|c|}{$F_{17}$} & \multicolumn{1}{c|}{0.543} & 0.000 & \multicolumn{1}{c|}{0.505($+$)} & 0.000 & \multicolumn{1}{c|}{0.485($+$)} & 0.000 & \multicolumn{1}{c|}{0.008($+$)} & 0.000 & \multicolumn{1}{c|}{0.703($-$)} & 0.000 & \multicolumn{1}{c|}{0.532($+$)} & 0.000 & \multicolumn{1}{c|}{\textbf{0.750($-$)}} & 0.000 \\ \hline
\multicolumn{1}{|c|}{$F_{18}$} & \multicolumn{1}{c|}{0.440} & 0.000 & \multicolumn{1}{c|}{0.170($+$)} & 0.000 & \multicolumn{1}{c|}{0.307($+$)} & 0.000 & \multicolumn{1}{c|}{0.177($+$)} & 0.000 & \multicolumn{1}{c|}{0.667($-$)} & 0.000 & \multicolumn{1}{c|}{\textbf{0.667($-$)}} & 0.000 & \multicolumn{1}{c|}{0.667($-$)} & 0.000 \\ \hline
\multicolumn{1}{|c|}{$F_{19}$} & \multicolumn{1}{c|}{0.375} & 0.000 & \multicolumn{1}{c|}{0.000($+$)} & 0.000 & \multicolumn{1}{c|}{0.023($+$)} & 0.000 & \multicolumn{1}{c|}{0.000($+$)} & 0.000 & \multicolumn{1}{c|}{0.503($-$)} & 0.000 & \multicolumn{1}{c|}{0.475($-$)} & 0.000 & \multicolumn{1}{c|}{\textbf{0.585($-$)}} & 0.000 \\ \hline
\multicolumn{1}{|c|}{$F_{20}$} & \multicolumn{1}{c|}{0.000} & 0.000 & \multicolumn{1}{c|}{0.025($-$)} & 0.000 & \multicolumn{1}{c|}{0.000($\approx$)} & 0.000 & \multicolumn{1}{c|}{0.000($\approx$)} & 0.000 & \multicolumn{1}{c|}{\textbf{0.483($-$)}} & 0.000 & \multicolumn{1}{c|}{0.275($-$)} & 0.000 & \multicolumn{1}{c|}{\textbf{0.483($-$)}} & 0.000 \\ \hline
\multicolumn{3}{|l|}{Total $+$   (Significantly better)} & \multicolumn{2}{c|}{5} & \multicolumn{2}{c|}{9} & \multicolumn{2}{c|}{12} & \multicolumn{2}{c|}{5} & \multicolumn{2}{c|}{7} & \multicolumn{2}{c|}{1} \\ \hline
\multicolumn{3}{|l|}{Total $\approx$ (Not   sig. different)} & \multicolumn{2}{c|}{7} & \multicolumn{2}{c|}{6} & \multicolumn{2}{c|}{6} & \multicolumn{2}{c|}{4} & \multicolumn{2}{c|}{6} & \multicolumn{2}{c|}{7} \\ \hline
\multicolumn{3}{|l|}{Total $-$   (Significantly worse)} & \multicolumn{2}{c|}{8} & \multicolumn{2}{c|}{5} & \multicolumn{2}{c|}{2} & \multicolumn{2}{c|}{11} & \multicolumn{2}{c|}{7} & \multicolumn{2}{c|}{12} \\ \hline
\multicolumn{1}{|c|}{\multirow{2}{*}{Func.}} & \multicolumn{2}{c|}{ADE-DDE} & \multicolumn{2}{c|}{ANDE} & \multicolumn{2}{c|}{NCD-DE} & \multicolumn{2}{c|}{CFNDE} & \multicolumn{2}{c|}{DIDE} & \multicolumn{2}{c|}{MOMMOP} & \multicolumn{2}{c|}{EMO-MMO} \\ \cline{2-15} 
\multicolumn{1}{|c|}{} & \multicolumn{1}{c|}{PR} & SR & \multicolumn{1}{c|}{PR} & SR & \multicolumn{1}{c|}{PR} & SR & \multicolumn{1}{c|}{PR} & SR & \multicolumn{1}{c|}{PR} & SR & \multicolumn{1}{c|}{PR} & SR & \multicolumn{1}{c|}{PR} & SR \\ \hline
\multicolumn{1}{|c|}{$F_{1}$} & \multicolumn{1}{c|}{\textbf{1.000($\approx$)}} & \textbf{1.000} & \multicolumn{1}{c|}{\textbf{1.000($\approx$)}} & \textbf{1.000} & \multicolumn{1}{c|}{\textbf{1.000($\approx$)}} & \textbf{1.000} & \multicolumn{1}{c|}{\textbf{1.000($\approx$)}} & \textbf{1.000} & \multicolumn{1}{c|}{\textbf{1.000($\approx$)}} & \textbf{1.000} & \multicolumn{1}{c|}{\textbf{1.000($\approx$)}} & \textbf{1.000} & \multicolumn{1}{c|}{\textbf{1.000($\approx$)}} & \textbf{1.000} \\ \hline
\multicolumn{1}{|c|}{$F_{2}$} & \multicolumn{1}{c|}{\textbf{1.000($\approx$)}} & \textbf{1.000} & \multicolumn{1}{c|}{\textbf{1.000($\approx$)}} & \textbf{1.000} & \multicolumn{1}{c|}{\textbf{1.000($\approx$)}} & \textbf{1.000} & \multicolumn{1}{c|}{\textbf{1.000($\approx$)}} & \textbf{1.000} & \multicolumn{1}{c|}{\textbf{1.000($\approx$)}} & \textbf{1.000} & \multicolumn{1}{c|}{\textbf{1.000($\approx$)}} & \textbf{1.000} & \multicolumn{1}{c|}{\textbf{1.000($\approx$)}} & \textbf{1.000} \\ \hline
\multicolumn{1}{|c|}{$F_{3}$} & \multicolumn{1}{c|}{\textbf{1.000($\approx$)}} & \textbf{1.000} & \multicolumn{1}{c|}{\textbf{1.000($\approx$)}} & \textbf{1.000} & \multicolumn{1}{c|}{\textbf{1.000($\approx$)}} & \textbf{1.000} & \multicolumn{1}{c|}{\textbf{1.000($\approx$)}} & \textbf{1.000} & \multicolumn{1}{c|}{\textbf{1.000($\approx$)}} & \textbf{1.000} & \multicolumn{1}{c|}{\textbf{1.000($\approx$)}} & \textbf{1.000} & \multicolumn{1}{c|}{\textbf{1.000($\approx$)}} & \textbf{1.000} \\ \hline
\multicolumn{1}{|c|}{$F_{4}$} & \multicolumn{1}{c|}{\textbf{1.000($\approx$)}} & \textbf{1.000} & \multicolumn{1}{c|}{\textbf{1.000($\approx$)}} & \textbf{1.000} & \multicolumn{1}{c|}{\textbf{1.000($\approx$)}} & \textbf{1.000} & \multicolumn{1}{c|}{\textbf{1.000($\approx$)}} & \textbf{1.000} & \multicolumn{1}{c|}{\textbf{1.000($\approx$)}} & \textbf{1.000} & \multicolumn{1}{c|}{0.985($+$)} & 0.940 & \multicolumn{1}{c|}{\textbf{1.000($\approx$)}} & \textbf{1.000} \\ \hline
\multicolumn{1}{|c|}{$F_{5}$} & \multicolumn{1}{c|}{\textbf{1.000($\approx$)}} & \textbf{1.000} & \multicolumn{1}{c|}{\textbf{1.000($\approx$)}} & \textbf{1.000} & \multicolumn{1}{c|}{\textbf{1.000($\approx$)}} & \textbf{1.000} & \multicolumn{1}{c|}{\textbf{1.000($\approx$)}} & \textbf{1.000} & \multicolumn{1}{c|}{\textbf{1.000($\approx$)}} & \textbf{1.000} & \multicolumn{1}{c|}{\textbf{1.000($\approx$)}} & \textbf{1.000} & \multicolumn{1}{c|}{\textbf{1.000($\approx$)}} & \textbf{1.000} \\ \hline
\multicolumn{1}{|c|}{$F_{6}$} & \multicolumn{1}{c|}{\textbf{1.000($-$)}} & \textbf{1.000} & \multicolumn{1}{c|}{\textbf{1.000($-$)}} & \textbf{1.000} & \multicolumn{1}{c|}{\textbf{1.000($-$)}} & \textbf{1.000} & \multicolumn{1}{c|}{\textbf{1.000($-$)}} & \textbf{1.000} & \multicolumn{1}{c|}{\textbf{1.000($-$)}} & \textbf{1.000} & \multicolumn{1}{c|}{\textbf{1.000($-$)}} & \textbf{1.000} & \multicolumn{1}{c|}{\textbf{1.000($-$)}} & \textbf{1.000} \\ \hline
\multicolumn{1}{|c|}{$F_{7}$} & \multicolumn{1}{c|}{0.838($-$)} & 0.039 & \multicolumn{1}{c|}{0.933($-$)} & 0.176 & \multicolumn{1}{c|}{0.913($-$)} & 0.078 & \multicolumn{1}{c|}{0.893($-$)} & 0.058 & \multicolumn{1}{c|}{0.921($-$)} & 0.040 & \multicolumn{1}{c|}{\textbf{1.000($-$)}} & \textbf{1.000} & \multicolumn{1}{c|}{\textbf{1.000($-$)}} & \textbf{1.000} \\ \hline
\multicolumn{1}{|c|}{$F_{8}$} & \multicolumn{1}{c|}{0.747($-$)} & 0.000 & \multicolumn{1}{c|}{0.944($-$)} & 0.078 & \multicolumn{1}{c|}{0.965($-$)} & 0.118 & \multicolumn{1}{c|}{0.984($-$)} & 0.784 & \multicolumn{1}{c|}{0.692($-$)} & 0.000 & \multicolumn{1}{c|}{\textbf{1.000($-$)}} & \textbf{1.000} & \multicolumn{1}{c|}{\textbf{1.000($-$)}} & \textbf{1.000} \\ \hline
\multicolumn{1}{|c|}{$F_{9}$} & \multicolumn{1}{c|}{0.384($-$)} & 0.000 & \multicolumn{1}{c|}{0.512($-$)} & 0.000 & \multicolumn{1}{c|}{0.572($-$)} & 0.000 & \multicolumn{1}{c|}{0.385($-$)} & 0.000 & \multicolumn{1}{c|}{0.571($-$)} & 0.000 & \multicolumn{1}{c|}{\textbf{1.000($-$)}} & \textbf{0.920} & \multicolumn{1}{c|}{0.950($-$)} & 0.000 \\ \hline
\multicolumn{1}{|c|}{$F_{10}$} & \multicolumn{1}{c|}{\textbf{1.000($\approx$)}} & \textbf{1.000} & \multicolumn{1}{c|}{\textbf{1.000($\approx$)}} & \textbf{1.000} & \multicolumn{1}{c|}{\textbf{1.000($\approx$)}} & \textbf{1.000} & \multicolumn{1}{c|}{\textbf{1.000($\approx$)}} & \textbf{1.000} & \multicolumn{1}{c|}{\textbf{1.000($\approx$)}} & \textbf{1.000} & \multicolumn{1}{c|}{\textbf{1.000($\approx$)}} & \textbf{1.000} & \multicolumn{1}{c|}{\textbf{1.000($\approx$)}} & \textbf{1.000} \\ \hline
\multicolumn{1}{|c|}{$F_{11}$} & \multicolumn{1}{c|}{\textbf{1.000($\approx$)}} & \textbf{1.000} & \multicolumn{1}{c|}{\textbf{1.000($\approx$)}} & \textbf{1.000} & \multicolumn{1}{c|}{\textbf{1.000($\approx$)}} & \textbf{1.000} & \multicolumn{1}{c|}{0.981($+$)} & 0.818 & \multicolumn{1}{c|}{\textbf{1.000($\approx$)}} & \textbf{1.000} & \multicolumn{1}{c|}{0.710($+$)} & 0.040 & \multicolumn{1}{c|}{\textbf{1.000($\approx$)}} & \textbf{1.000} \\ \hline
\multicolumn{1}{|c|}{$F_{12}$} & \multicolumn{1}{c|}{\textbf{1.000($-$)}} & \textbf{1.000} & \multicolumn{1}{c|}{\textbf{1.000($-$)}} & \textbf{1.000} & \multicolumn{1}{c|}{0.993($-$)} & 0.941 & \multicolumn{1}{c|}{0.977($-$)} & 0.727 & \multicolumn{1}{c|}{\textbf{1.000($-$)}} & \textbf{1.000} & \multicolumn{1}{c|}{0.955($-$)} & 0.660 & \multicolumn{1}{c|}{\textbf{1.000($-$)}} & \textbf{1.000} \\ \hline
\multicolumn{1}{|c|}{$F_{13}$} & \multicolumn{1}{c|}{0.686($+$)} & 0.000 & \multicolumn{1}{c|}{0.686($+$)} & 0.000 & \multicolumn{1}{c|}{0.902($+$)} & 0.471 & \multicolumn{1}{c|}{0.774($+$)} & 0.000 & \multicolumn{1}{c|}{0.987($+$)} & 0.920 & \multicolumn{1}{c|}{0.667($+$)} & 0.000 & \multicolumn{1}{c|}{0.997($-$)} & 0.980 \\ \hline
\multicolumn{1}{|c|}{$F_{14}$} & \multicolumn{1}{c|}{0.667($+$)} & 0.000 & \multicolumn{1}{c|}{0.667($+$)} & 0.000 & \multicolumn{1}{c|}{0.683($+$)} & 0.000 & \multicolumn{1}{c|}{0.667($+$)} & 0.000 & \multicolumn{1}{c|}{0.773($+$)} & 0.020 & \multicolumn{1}{c|}{0.667($+$)} & 0.000 & \multicolumn{1}{c|}{0.733($+$)} & 0.060 \\ \hline
\multicolumn{1}{|c|}{$F_{15}$} & \multicolumn{1}{c|}{0.637($+$)} & 0.000 & \multicolumn{1}{c|}{0.632($+$)} & 0.000 & \multicolumn{1}{c|}{0.654($+$)} & 0.000 & \multicolumn{1}{c|}{\textbf{0.750($-$)}} & 0.000 & \multicolumn{1}{c|}{0.748($-$)} & 0.000 & \multicolumn{1}{c|}{0.618($+$)} & 0.000 & \multicolumn{1}{c|}{0.595($+$)} & 0.000 \\ \hline
\multicolumn{1}{|c|}{$F_{16}$} & \multicolumn{1}{c|}{0.667($+$)} & 0.000 & \multicolumn{1}{c|}{0.667($+$)} & 0.000 & \multicolumn{1}{c|}{0.667($+$)} & 0.000 & \multicolumn{1}{c|}{0.667($+$)} & 0.000 & \multicolumn{1}{c|}{0.667($+$)} & 0.000 & \multicolumn{1}{c|}{0.630($+$)} & 0.000 & \multicolumn{1}{c|}{0.657($+$)} & 0.000 \\ \hline
\multicolumn{1}{|c|}{$F_{17}$} & \multicolumn{1}{c|}{0.375($+$)} & 0.000 & \multicolumn{1}{c|}{0.397($+$)} & 0.000 & \multicolumn{1}{c|}{0.522($+$)} & 0.000 & \multicolumn{1}{c|}{0.545($-$)} & 0.000 & \multicolumn{1}{c|}{0.593($-$)} & 0.000 & \multicolumn{1}{c|}{0.505($+$)} & 0.000 & \multicolumn{1}{c|}{0.335($+$)} & 0.000 \\ \hline
\multicolumn{1}{|c|}{$F_{18}$} & \multicolumn{1}{c|}{0.654($-$)} & 0.000 & \multicolumn{1}{c|}{0.654($-$)} & 0.000 & \multicolumn{1}{c|}{\textbf{0.667($-$)}} & 0.000 & \multicolumn{1}{c|}{\textbf{0.667($-$)}} & 0.000 & \multicolumn{1}{c|}{\textbf{0.667($\approx$)}} & 0.000 & \multicolumn{1}{c|}{0.497($-$)} & 0.000 & \multicolumn{1}{c|}{0.327($+$)} & 0.000 \\ \hline
\multicolumn{1}{|c|}{$F_{19}$} & \multicolumn{1}{c|}{0.375($\approx$)} & 0.000 & \multicolumn{1}{c|}{0.363($+$)} & 0.000 & \multicolumn{1}{c|}{0.505($-$)} & 0.000 & \multicolumn{1}{c|}{0.505($-$)} & 0.000 & \multicolumn{1}{c|}{0.543($-$)} & 0.000 & \multicolumn{1}{c|}{0.230($+$)} & 0.000 & \multicolumn{1}{c|}{0.135($+$)} & 0.000 \\ \hline
\multicolumn{1}{|c|}{$F_{20}$} & \multicolumn{1}{c|}{0.250($-$)} & 0.000 & \multicolumn{1}{c|}{0.248($-$)} & 0.000 & \multicolumn{1}{c|}{0.255($-$)} & 0.000 & \multicolumn{1}{c|}{0.302($-$)} & 0.000 & \multicolumn{1}{c|}{0.355($-$)} & 0.000 & \multicolumn{1}{c|}{0.125($-$)} & 0.000 & \multicolumn{1}{c|}{0.080($-$)} & 0.000 \\ \hline
\multicolumn{1}{|c|}{$+$} & \multicolumn{2}{c|}{5} & \multicolumn{2}{c|}{6} & \multicolumn{2}{c|}{5} & \multicolumn{2}{c|}{4} & \multicolumn{2}{c|}{3} & \multicolumn{2}{c|}{8} & \multicolumn{2}{c|}{6} \\ \hline
\multicolumn{1}{|c|}{$\approx$} & \multicolumn{2}{c|}{8} & \multicolumn{2}{c|}{7} & \multicolumn{2}{c|}{7} & \multicolumn{2}{c|}{6} & \multicolumn{2}{c|}{8} & \multicolumn{2}{c|}{5} & \multicolumn{2}{c|}{7} \\ \hline
\multicolumn{1}{|c|}{$-$} & \multicolumn{2}{c|}{7} & \multicolumn{2}{c|}{7} & \multicolumn{2}{c|}{8} & \multicolumn{2}{c|}{10} & \multicolumn{2}{c|}{9} & \multicolumn{2}{c|}{7} & \multicolumn{2}{c|}{7} \\ \hline
\end{tabular}%
}
\end{table*}

\subsection{State-of-the-Art Algorithms for Comparison} \label{sec:expG_results}

We compared E-k-BBBC with thirteen other MMOEAs from the state-of-the-art:
SPEA-DE/rand/1~\cite{xia2024learning},
SSGA~\cite{de2014dynamic},
CDE~\cite{thomsen2004multimodal},
RS-CMSA~\cite{ahrari2017multimodal},
LBP-ADE~\cite{zhao2019local},
HillVallEA18~\cite{maree2018real},
ADE-DDE~\cite{wang2020guardhealth},
ANDE~\cite{wang2019automatic},
NCD-DE~\cite{jiang2021optimizing},
CFNDE~\cite{ma2023coarse},
DIDE~\cite{chen2019distributed},
MOMMOP~\cite{wang2014mommop},
EMO-MMO~\cite{cheng2017evolutionary}	
-- most of these algorithms participated in competitions for either CEC or GECCO throughout 2017 and 2020 (the latest at the time of this work), others are taken from the latest published articles in the literature. However, we could not formally compare E-k-BBBC to RS-CMSA-EAII~\cite{ahrari2021static}, one of the most efficient multimodal optimizers, since the original paper does not report the values for SR. 

\subsection{Parameter Settings}
The settings of E-k-BBBC were set accordingly to Sec.~\ref{sec:settings}; however, some problems have an unknown and large number of local optima (i.e., $F_6$, $F_8$, $F_{11}$--$F_{20}$), making impossible to set the value of $m$. Therefore, we set $m$ equal to the number of global optima, knowing that E-k-BBBC will converge to a random local optima and potentially miss the requested peaks.

On the other hand, the results of the state-of-the-art algorithms were collected from the CEC repository or according to their corresponding references. Note that these methods were purposely developed to locate global peaks, contrary to E-k-BBBC. Therefore, we expected E-k-BBBC to be outperformed by most of these methods in this context.

\subsection{Results}

We conducted fifty independent runs (NR$=50$) and averaged the outcomes for a fair comparison. To statistically assess the results, we applied the Wilcoxon rank-sum test~\cite{demvsar2006statistical}, using a significance level of $\alpha=0.05$. In the result table, a $+$ sign indicates when E-k-BBBC significantly outperforms, a $-$ sign marks when it performs significantly worse, and a $\approx$ sign denotes no significant difference.

Table~\ref{tab:comparison_4} presents the experimental results of PR and SR on the CEC'2013 benchmark set with an accuracy level of $\varepsilon = \expnumber{1.0}{-04}$. The displayed values represent the best outcomes obtained by E-k-BBBC and other algorithms, with the top PR and SR for each function highlighted in bold. As expected, E-k-BBBC achieved the best performance on only eight out of twenty functions. Nevertheless, it outperforms three algorithms (SSGA, CDE, and MOMMOP) and performed competitively against six (SPEA-DE/rand/1, LBP-ADE, ADE-DDE, ANDE, NCD-DE, EMO-MMO). 

This is due to the vast number of local optima, which were not included in the setting of the parameter $m$, showing k-BBBC's biggest limitation.
This is particularly evident in the two functions that scored NP $=0.000$: 
\begin{itemize}
    \item $F_{8}$, the Shubert function with $d=3$ dimensions, features a considerable number of local optima, trapping k-BBBC away from global ones -- i.e., the probability of finding a global optimum by limiting the value of $m$ (and therefore, $k$) is almost zero; and 
    \item $F_{20}$, a composite function with $d=20$ dimensions which, according to the results in Sec.~\ref{sec:high_dim_experiments}, should have been easily solved -- although this is notably the hardest function in CEC'2013 that only few state-of-the-art algorithms can tackle (NP $=0.483$ with RSCMSA~\cite{ahrari2017multimodal} and HillVallEA18~\cite{maree2018real}, and a remarkable NP $=0.618$ with RS-CMSA-EAII~\cite{ahrari2021static}). 
\end{itemize}
However, even with this limitation, k-BBBC scored the highest value on $F_{16}$, proving that applying clustering to the basic BBBC algorithm is a promising direction to solve multimodal optimization problems. although being third after RS-CMSA-EAII with NP $=0.833$~\cite{ahrari2021static} and HillVallEA19 with NP $=0.723$~\cite{maree2019benchmarking}.

\section{Discussion}
\label{sec:discussion}

Based on the results reported in Sec.~\ref{sec:experiments_local} and ~\ref{sec:experiments_global}, we can summarize several advantages and disadvantages of k-BBBC.

\subsection{Advantages}
\label{sec:advantages}

We report a list of the advantages of k-BBBC:

\begin{itemize}
    \item The algorithm features high accuracy in both search space and objective space while retrieving (almost) all the local optima for most of the problems (when possible) -- regardless of their dimensionality. This is especially evident while using elitism.
    \item In most cases, k-BBBC outperforms MCS, FSDE, CP, EPS, CSA, AIN, SSGA, CDE, and MOMMOP in terms of accuracy and success rate, and performs competitively against SPEA-DE/rand/1, LBP-ADE, ADE-DDE, ANDE, NCD-DE, EMO-MMO.
    \item The entire population converges to the local optima, not only the best individuals. This is due to the denominator term in Eqn.~(\ref{eq:bang}), which reduces the range of exploration of new individuals from their center of mass -- which could be modified to fasten the convergence and increase the accuracy. The range of exploration is decreased at each generation of k-BBBC (and equivalently to its original version for global optimization), guaranteeing convergence in the entire population. Having the entire population converging to the local optima allows for easy identification of the best individuals, and inspired us to develop the procedures described in Sec.~\ref{sec:postprocessing}. 
    \item Directly from the previous advantage, using the quantification method (Algorithm~\ref{alg:missing_quantification}) allows us to have an estimation of how the algorithm performed in terms of success rate without the need of knowing the exact value of the actual optima -- which we will not have because the purpose of a multimodal optimization algorithm is to find those values. Based on the results summarized in Sec.~\ref{sec:results}, where we compared the actual success rate with the one obtained by quantification, we can claim that the quantification method works properly when the population converges to local optima. This, however, can be exploited only if the number of optima $m$ is known or can be estimated. 
    \item The algorithm can tackle high-dimensional problems with full retrieval of all local optima (see Table~\ref{tab:results_high}), whereas other algorithms failed (MCS, FSDE, CP, EPS, CSA, AIN, and CDE). Additionally, when compared with the work by Deb and Saha~\cite{deb2012multimodal}, their bi-objective NSGA-II-based algorithm was not tested for problems with $d>16$, whereas k-BBBC works well also for $d=32$. 
    \item k-BBBC works with non-differentiable functions as well, whereas, the algorithm of Deb and Saha~\cite{deb2012multimodal} uses the derivative of the objective function as the second objective to be minimized. This is a limitation that k-BBBC does not have.
    \item k-BBBC works for both minimization and maximization problems.
    \item k-BBBC inherits the advantages of the original single-modal global optimizer BBBC~\cite{erol2006new}: fast convergence, good balance between exploration and exploitation, and fast execution time (see Sec.~\ref{sec:complexity}). Considering k-BBBC performance, using BBBC as a base for multimodal optimization is a promising path that is worth investigating further -- more discussion on future works is provided in Sec.~\ref{sec:conclusion}.
\end{itemize}

\subsection{Disadvantages}
\label{sec:disadvantages}

We report a list of the disadvantages of k-BBBC:

\begin{itemize}
    \item It is required to know the number of local optima of a problem or to have a valid estimation of it. This might not always be possible (especially as shown in the problems from the CEC'2013 benchmark set, Sec.~\ref{sec:experiments_global}); however, in some cases, this number can be empirically retrieved: one could get a population that correctly converges to all local optima; then, use the silhouette or even manually count the number of optima, and use Algorithm~\ref{alg:identification} to find out the actual number.
    \item It relies on a mathematical estimation of the number of clusters for correct retrieval. Although the formulation reported in Eqn.~(\ref{eq:k}) and (\ref{eq:n}) are appropriate for the problems tested in this study, it is not a guarantee of full reliability.  
    \item It is based on clustering and, therefore, relies on algorithms such as k-means and k-medoids, with all their related issues and limitations (e.g., see Figure~\ref{fig:kmeans_error}). Furthermore, when running k-BBBC on the problem with $d=32$, we observed that k-means failed to converge in 200 iterations (a limit we empirically set based on the problems we tested). This did not prevent k-BBBC from retrieving every optimum, but it is important to report this issue -- to the best of our knowledge, there is no hard limit for the dimensionality that k-means can tackle; however, it is reasonable to assume that the higher the dimensionality, the less the convergence.
    \item Problems featuring non-uniform distribution of local optima in the search space may deceive the clustering algorithm, especially in regions with a high density of local optima -- i.e., k-means can group all those solutions in the same cluster. This limitation comes directly from the need to know the number of optima to be sought and must be addressed to improve the results of k-BBBC with additional niching methods.
    \item Based on the analysis reported in Sec.~\ref{sec:complexity}, the algorithm's runtime increases quadratically with the number of local optima of the problem and cubically with its dimensionality. Therefore, problems featuring high dimensionality are expected to have a higher runtime -- e.g., as shown in Table~\ref{tab:results_high}, a single run for $d=32$ took approximately eleven hours. However, we can consider optimization -- in most cases -- as an offline process, and this timing still falls within an acceptable range.
    \item The algorithm is not meant for global peaks only, as it does not distinguish a global peak from a local peak, or even a plateau. This could be addressed by implementing niching methods and/or survival strategies.
    
\end{itemize}

\section{Conclusion}
\label{sec:conclusion}

In this paper, we presented k-BBBC, an MMEA based on clustering. Additionally, we introduce a post-processing procedure to identify local optima in a population of solutions and quantify how many have been missed by the algorithm without the need to compare them with the actual points. We compared k-BBBC with other algorithms in the literature, and our results highlight high accuracy and, particularly, a higher success rate in retrieving optima. Additionally, k-BBBC can tackle problems with a high number of optima (tested on $m=379$) and high dimensionalities (tested on $d=32$). 

However, k-BBBC requires an insight into the optima number, as it relies on k-means. This limitation caused low performance on the CEC'2013 benchmark set when compared to other specialized MMEAs, which focus on niching. Interestingly, k-BBBC (particularly its elitist version) could compete with nine out of thirteen state-of-the-art algorithms developed in the last decade. This indicates a promising path to be explored in multimodal optimization problems, based on clustering and the BBBC algorithm. We believe that, when extended with specific niching operators and a non-parametric clustering method, the BBBC algorithm could achieve top performance without falling into the trap of local optima -- which was intended for this work. Future works will investigate this, especially by removing the burden of defining the number of expected optima. Lastly, we plan to test the algorithm on real-engineering applications, such as retrieving alternative sub-optimal designs for robotic devices~\cite{stroppa2023optimizing, stroppa2024optimizing}.

\section*{Code}
The MATLAB script is available on EVO Lab’s MathWorks File Exchange repository: \url{www.mathworks.com/matlabcentral/fileexchange/157081}.

\section*{Acknowledgments}

This work is funded by TUB\.ITAK within the scope of the 2232-B International Fellowship for Early Stage Researchers Program number 121C145.

 \bibliographystyle{elsarticle-num} 
 \bibliography{references}

\end{document}